\documentclass{article}

\PassOptionsToPackage{most}{tcolorbox}
\PassOptionsToPackage{hidelinks}{hyperref}
\usepackage[final]{acl}

\usepackage{times}
\usepackage{latexsym} 
\usepackage[utf8]{inputenc}
\usepackage[T1]{fontenc}
\usepackage{amsmath,amssymb}
\usepackage{amsfonts}
\usepackage{booktabs}
\usepackage{microtype}
\usepackage{inconsolata}
\usepackage{graphicx}
\usepackage{wrapfig}
\usepackage{enumitem}
\usepackage{pifont}
\usepackage{nicefrac}
\usepackage{listings}
\usepackage{fontawesome5}
\makeatletter
\@ifpackageloaded{hyperref}{}{\usepackage{hyperref}}
\makeatother

\tcbuselibrary{listings, breakable, skins}
\definecolor{DeepPurple}{HTML}{673AB7}
\definecolor{LighterGray}{HTML}{FAFAFA}
\definecolor{CaseOrange}{HTML}{F57C00}

\newcommand{\answerTODO}[1][]{\textcolor{red}{\bf [TODO]}}
\newcommand{\justificationTODO}[1][]{\textcolor{red}{\bf [TODO]}}

\title{SkillGenBench: Benchmarking \\Skill Generation Pipelines for LLM Agents
}

\author{%
  Yifan Zhou$^{1*}$\quad
  Zhentao Zhang$^{2*}$\quad
  Ziming Cheng$^{3*}$\quad
  Shuo Zhang$^{4*}$\\
  \textbf{Qizhen Lan}$^{4}$\quad
  \textbf{Zhangquan Chen}$^{5}$\quad
  \textbf{Zhi Yang}$^{6}$\quad
  \textbf{Qianyu Xu}$^{7}$\\
  \textbf{Ronghao Chen}$^{4,8\dagger}$\quad
  \textbf{Huacan Wang}$^{4,9\dagger}$\quad
  \textbf{Sen Hu}$^{4,8\dagger}$\\[6pt]
  $^{1}$SJTU\quad
  $^{2}$XJTU\quad
  $^{3}$NUS\quad
  $^{4}$QuantaAlpha\quad
  $^{5}$THU\quad
  $^{6}$SUFE\quad
  $^{7}$NTU\quad
  $^{8}$PKU\quad
  $^{9}$UCAS\\
  {\small $^{*}$Equal Contribution\quad $^{\dagger}$Correspondence: chenronghao@alumni.pku.edu.cn  wanghuacan17@mails.ucas.ac.cn  husen@pku.edu.cn}\\[6pt]
  {\small \href{https://github.com/QuantaAlpha/SkillGenBench}{\textcolor{DeepPurple}{\faIcon{github}\ \nolinkurl{https://github.com/QuantaAlpha/SkillGenBench}}}}
}

\begin{document}

\maketitle

\begin{abstract}
As LLM agents are increasingly built around reusable \emph{skills}, a central challenge is no longer only whether agents can \emph{use} provided skills, but whether they can \emph{generate} correct, reusable, and executable skills from repositories and documents. Existing benchmarks primarily evaluate the efficacy of given skills or the ability of agents to solve downstream tasks from raw context, but they do not isolate \emph{skill generation} itself as the object of study. We introduce \textbf{SkillGenBench}, a benchmark for evaluating skill generation pipelines under a unified and controlled protocol. In SkillGenBench, a generator receives raw corpora and produces standardized skill artifacts, which are then executed under fixed harnesses and assessed with unified evaluation procedures. The benchmark covers two generation regimes: \emph{task-conditioned generation}, where a task-specific skill is synthesized after the task is revealed, and \emph{task-agnostic generation}, where a reusable skill library must be distilled before downstream tasks are known. It also spans two complementary procedural sources: \emph{repository-grounded} instances, where procedures are distributed across code, configuration, and scripts, and \emph{document-grounded} instances, where procedures and constraints must be distilled from long-form text. We provide standardized task specifications, pinned environments, and evaluation protocols centered on deterministic execution-based checks, supplemented by auxiliary signals for diagnosis. Experiments across a range of skill-generation methods and backbones show substantial performance variation, highlight the difficulty of reusable skill distillation, and reveal distinct failure modes in skill generation from software repositories versus long-form documents.  SkillGenBench establishes a reproducible testbed for studying skill generation as an independent research problem in agent systems.
\end{abstract}

\section{Introduction}
\label{sec:intro}

\begin{figure}[t]
  \centering
  \includegraphics[width=\linewidth,height=0.33\textheight,keepaspectratio]{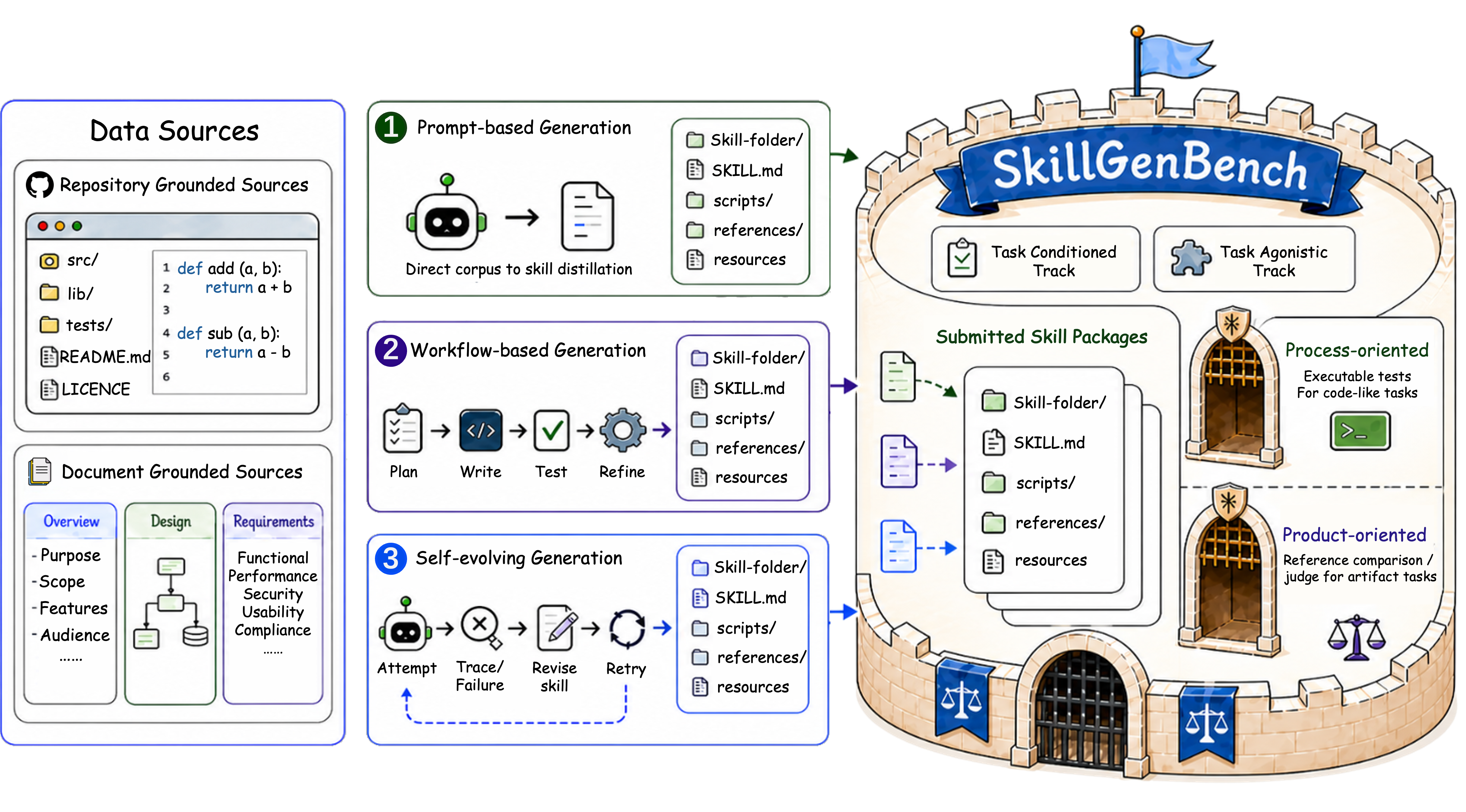}
  \caption{Overview of SkillGenBench. Skill-generation pipelines transform repository- and document-grounded sources into standardized skill packages, which are evaluated under task-conditioned and task-agnostic tracks with fixed execution checks and artifact-level diagnostics.}
  \label{fig:skillgenbench_overview}
\end{figure}


As LLM agents are deployed in increasingly complex environments, a growing design trend is to move beyond monolithic prompting and toward modular, persistent capability abstractions. Recent agent systems increasingly rely not only on runtime augmentation through tools and external context, but also on reusable \emph{skills}: packaged procedural artifacts that encode how to accomplish classes of tasks in a form that can be stored, versioned, and reused. In emerging Agent Skills interfaces~\cite{anthropic2025agentskills}, a skill is typically organized around a \texttt{SKILL.md} file together with optional scripts, references, and auxiliary resources. This packaging abstraction offers practical advantages that raw in-context reasoning does not naturally provide: skills can be audited, cached, shared across agents and teams, updated independently of the base model, and composed into larger workflows. As a result, skills are becoming an increasingly important substrate for scalable agent development.


Early empirical evidence highlights both the promise and the fragility of skill-based agent design. 
\textit{CL-Bench}~\cite{clbench} shows that even when relevant evidence is explicitly present in complex context, models frequently fail to extract and operationalize it into correct procedures.
In parallel, \textit{SkillsBench}~\cite{skillsbench2026} shows that curated skills can substantially improve downstream task performance, while automatically generated skills—especially those produced on the fly—are often unstable and can even induce negative transfer.
Taken together, these findings suggest a broader lesson: procedural knowledge is valuable when externalized into structured artifacts, but difficult for models to reliably distill from raw repositories, documents, and other unstructured corpora.


This tension becomes more important in realistic deployment settings, where procedural knowledge is not static. New repositories, APIs, technical documents, and papers continuously introduce new constraints, workflows, and best practices that must be incorporated if agents are to remain current~\citep{skillnet}. In such settings, the central challenge is not only whether an agent can \emph{use} a provided skill, but whether a pipeline can \emph{generate} a correct, reusable, and executable skill from visible corpora. Yet existing benchmarks rarely isolate this generation step as the primary object of evaluation. Skill-centric benchmarks~\citep{skillsbench2026,skillusagewild} typically measure whether a provided skill improves downstream execution; task-centric benchmarks~\citep{agentbench,webarena,swebench,terminalbench,clbench} measure whether an agent can solve an end task from raw context. Neither offers a controlled protocol for comparing \emph{skill generation pipelines} as modular, interchangeable components under fixed downstream execution conditions.


We address this gap with \textbf{SkillGenBench}, a benchmark for evaluating \textbf{skill generation pipelines} under a unified and controlled protocol. 
As showed in Figure~\ref{fig:skillgenbench_overview}, SkillGenBench treats the generator itself as the object of study: given raw corpora, a generator produces standardized skill artifacts, which are then executed under fixed harnesses and assessed with a unified evaluation procedure. We study two practically important generation regimes. In \textbf{task-conditioned generation}, the generator synthesizes a task-specific skill from a raw corpus together with the task specification. In \textbf{task-agnostic generation}, the generator must distill a reusable skill library from raw corpora \emph{before} downstream tasks are revealed. The resulting library is generated once and then reused without regeneration, allowing us to evaluate not only task-level effectiveness but also abstraction quality, compression, and cross-task reuse.


SkillGenBench spans two complementary procedural sources. \textbf{Repository-grounded} instances require generators to recover procedures distributed across repository structure, code, configuration, and scripts. \textbf{Document-grounded} instances require generators to distill procedures and constraints from long-form knowledge sources whose relevant evidence may be explicit but dispersed. To enable reproducible comparison, we provide standardized task specifications, pinned environments, fixed execution harnesses, and unified evaluation protocols centered on deterministic execution-based checks, supplemented by auxiliary similarity-based and judge-based signals where needed for diagnosis~\citep{skillsbench2026,swebench,terminalbench}. 
Our contributions are: (1) a benchmark that directly evaluates skill generation pipelines, rather than provided skills or unconstrained end-to-end agents; (2) a task-agnostic setting that measures one-shot reusable skill library distillation before hidden downstream tasks are revealed; (3) a unified benchmark spanning both repository-grounded and document-grounded procedural knowledge; and (4) a reproducible empirical study across representative generator families with systematic failure analysis.

\begin{table}[t]
  \caption{Comparison with representative skill-related and agent benchmarks.}
  \label{tab:benchmark_comparison}
  \centering
  \small
  \setlength{\tabcolsep}{3.5pt}
  \renewcommand{\arraystretch}{1.0}

  \begin{tabular*}{\linewidth}{@{\extracolsep{\fill}} lcccc @{}}
    \toprule
    \textbf{Dimension}
      & \textbf{SkillGenBench}
      & \textbf{SkillsBench}
      & \textbf{SWE-Skills}
      & \textbf{WildClawBench} \\
    \midrule
    Primary target
      & Skill gen.
      & Skill efficacy
      & SWE skills
      & Agent ability \\
    
    Skill acquisition
      & Self-generated
      & Curated/self-gen.
      & Public SWE skills
      & Runtime capability \\
    
    Task-conditioned generation
      & $\checkmark$
      & $\checkmark$
      & $\times$
      & $\times$ \\
    
    Task-agnostic generation
      & $\checkmark$
      & $\times$
      & $\times$
      & $\times$ \\
    
    Process-oriented evaluation
      & $\checkmark$
      & $\checkmark$
      & $\checkmark$
      & $\checkmark$ \\
    
    Product-oriented comparison
      & $\checkmark$
      & $\times$
      & $\times$
      & $\times$ \\
    \bottomrule
  \end{tabular*}
\end{table}


\section{Related Work}
\label{sec:related}

\subsection{Agent Skills and Runtime Augmentation}

A substantial body of work extends agent capability through runtime augmentation, including reasoning-and-acting loops~\citep{react}, tool use~\citep{toolformer,toollm}, retrieval augmentation~\citep{rag}, and standardized interfaces such as MCP~\citep{mcp}. While effective, these approaches primarily improve what an agent can accomplish within a single execution episode, leaving procedural knowledge implicitly embedded in prompts, traces, or retrieved context.

Recent work increasingly treats \emph{skills} as reusable procedural artifacts that persist beyond individual executions~\citep{sokagentskills}. Early systems acquire or consolidate skills from agent experience without standardized packaging~\citep{voyager,reflexion,expel,cascade}, whereas recent frameworks adopt explicit skill abstractions with portable packaging interfaces~\citep{anthropic2025agentskills}. Building on this abstraction, subsequent work studies skill creation~\citep{skillnet}, orchestration and routing~\citep{agentskillos,skillrouter}, and reusable skill ecosystems for long-horizon agent workflows.

\subsection{Skill Generation Pipelines}

As skills are increasingly treated as first-class artifacts, \emph{skill generation} has emerged as an important research direction that distills procedural knowledge from repositories, documentation, papers, and agent experience into reusable skill artifacts. Existing methods broadly follow three patterns: extracting and distilling procedures from corpora into structured skills~\cite{skillnet}, experience-driven consolidation from successful interactions or trajectories~\citep{skillx,autoskill,trace2skill}, and iterative refinement through execution feedback or structural validation~\citep{skillweaver,evoskill,skillrl,skillclaw,contractskill,mementoskills}. However, these generators are typically evaluated together with bespoke executors, routing policies, retrieval configurations, and environment assumptions. This coupling makes it difficult to disentangle the quality of skill generation from downstream integration choices. As a result, the field still lacks a benchmark that compares skill generation pipelines themselves as modular and interchangeable components under a common downstream protocol.

\subsection{Skill Benchmarks}

Recent skill benchmarks primarily evaluate \emph{skill efficacy}, assessing whether a provided skill artifact improves downstream execution for a given task under controlled evaluation settings~\citep{skillsbench2026,skillusagewild}. SkillsBench~\cite{skillsbench2026} compares the settings of no-skill, curated skill, and self-generated skills in domains under deterministic verifiers, and shows that curated skills can be beneficial while self-generated skills can be unstable. SWE-Skills-Bench~\cite{sweskillsbench} applies the same paired-evaluation logic to software engineering by pairing collected public skills with real-world repositories pinned at fixed commits with requirement-driven, execution-based verification. Although valuable, these benchmarks do not systematically evaluate \emph{skill generation pipelines}. 
As summarized in Table~\ref{tab:benchmark_comparison}, existing benchmarks differ in domain and execution environment, but they do not cover a task-agnostic library setting in which reusable skills must be distilled before hidden tasks are revealed. In SkillsBench, the self-generated condition is task-local: skills are generated only after the task is revealed and consumed immediately, rather than being compared under a common protocol across dedicated generators. SkillGenBench complements this line of work by treating skill generation pipelines as the primary object of evaluation, comparing pipelines that transform visible corpora into reusable skill artifacts and testing those artifacts under fixed harnesses and deterministic verification.




\section{SkillGenBench}
\label{sec:benchmark}

\begin{figure}
    \centering
    \includegraphics[width=\linewidth]{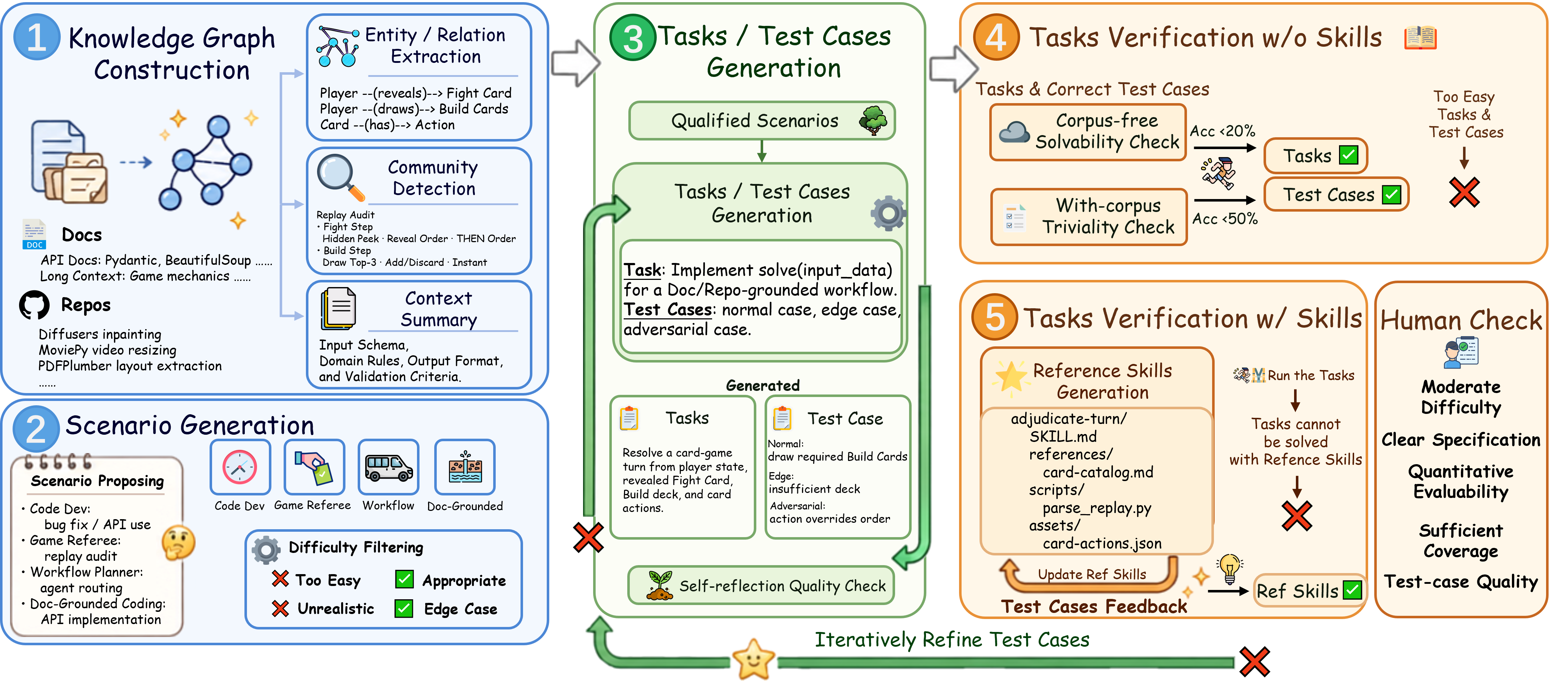}
    \caption{SkillGenBench construction pipeline. Repositories and long documents are first abstracted into a knowledge graph (Stage 1). Task scenarios are then proposed and filtered (Stage 2), and each scenario produces tasks and their test cases (Stage 3). Stage 4 filters out tasks solvable without procedural extraction or trivially solvable with the full corpus, and Stage 5 validates the remaining tasks with an iteratively refined reference skill. Tasks failing Stage 4 or Stage 5 are returned to Stage 3 for test-case refinement. Accepted tasks finally undergo human verification.}
    
    \label{fig:benchmark_construction}
\end{figure}

SkillGenBench evaluates how well LLMs can distill deployable, reusable skills from complex source materials and apply them to downstream tasks. Unlike benchmarks that assess the end-to-end task solving of agents~\citep{agentbench,webarena,swebench,terminalbench}, SkillGenBench treats skill generation itself as the primary object of evaluation. The agent first analyzes the source materials and generates a skill; a separate executor then invokes that skill to complete downstream tasks. By decoupling skill generation from execution, SkillGenBench provides a more direct measure of \emph{procedure-to-skill distillation}, rather than conflating it with downstream agentic capabilities such as task interpretation, planning, and tool use. 

At the instance level, each benchmark item is packaged as a containerized environment comprising five components: source materials, task specification, skill interface, executor, and evaluation protocol.

\begin{wrapfigure}{r}{0.50\linewidth}
    \vspace{-1.2em}
    \centering
    \includegraphics[width=\linewidth]{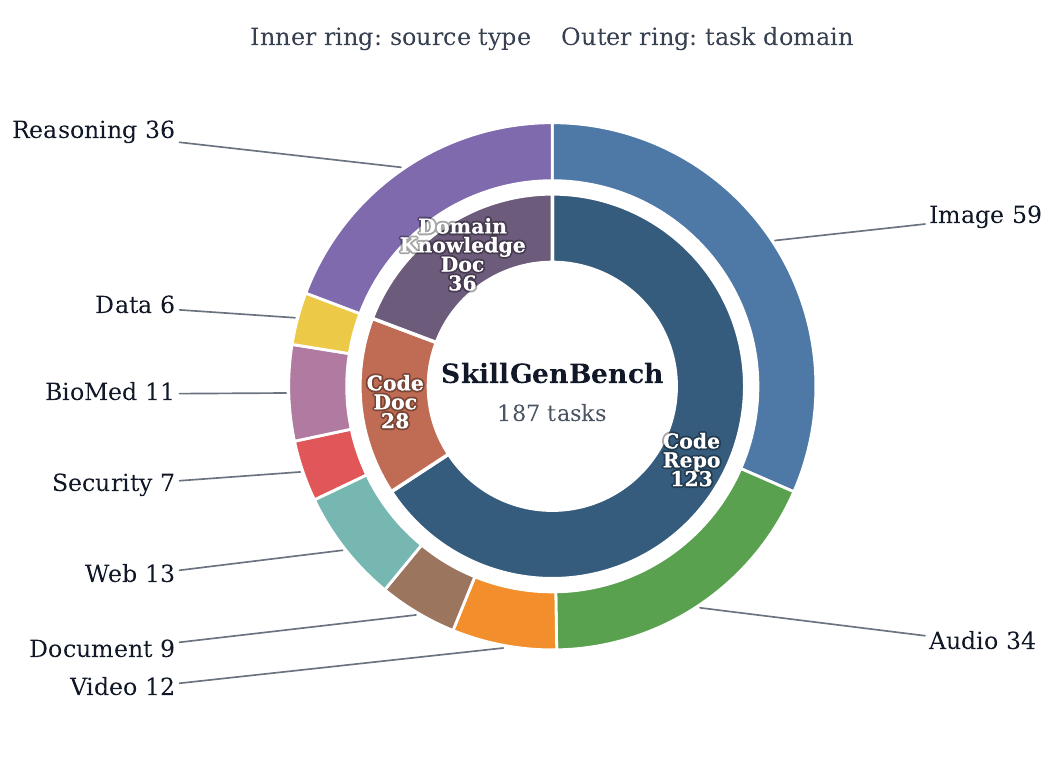}
    \caption{Source and domain composition of SkillGenBench. The inner ring shows source types and the outer ring shows task domains.}
    \label{fig:task_domain_overview}
    \vspace{-0.2in}
\end{wrapfigure}

\subsection{Sources of Procedural Knowledge}

SkillGenBench instances are organized by the source of procedural knowledge: \emph{repository-grounded} and \emph{document-grounded}. The two differ in how procedures are presented in the source materials and, accordingly, in what the model must extract.

\paragraph{Repository-grounded instances.}
The source materials consist of a code repository snapshot, including directory structures, README files, configuration files, dependency scripts, and environment conventions. Procedural knowledge is rarely stated explicitly; instead, it is implicit in code organization, call relations, entry scripts, and runtime constraints. The model must recover these latent workflows from the repository and distill them into a reusable skill.

\paragraph{Document-grounded instances.}
The source materials consist of dense, long-form texts such as system manuals, API specifications, and technical reports. In contrast to repository-grounded instances, procedural knowledge is expressed explicitly but distributed across passages, taking forms such as conditional branches, parameter rules, prerequisites, and ordered steps. The model must integrate these scattered constraints into a single skill that can be invoked on downstream tasks.

In the released benchmark, document-grounded instances are further separated into code documentation and domain-knowledge documentation subsets for analysis. Code documentation tasks emphasize API and library semantics, whereas domain-knowledge documentation tasks emphasize rule application and exact output constraints. Figure~\ref{fig:task_domain_overview} summarizes the 187-task benchmark composition across source types and task domains.

\subsection{Skill Generation Settings}

SkillGenBench defines two task settings based on whether the downstream task is known at generation time. The \emph{task-conditioned} setting reveals the task to the model; the \emph{task-agnostic} setting does not.

\paragraph{Task-conditioned setting.}
The model receives the source materials together with a task specification, and must identify the procedures most relevant to the task and distill them into a focused skill. This setting evaluates targeted distillation: whether the model can filter out irrelevant information and recover the key procedure required by the task. 

\paragraph{Task-agnostic setting.}
The model receives only the source materials, with no access to downstream tasks. It must build a reusable skill library within a fixed generation budget; this library is then used to support held-out tasks revealed at execution time. The challenge here is not task-specific synthesis but the identification of procedures with cross-task reuse value, and their organization into deployable skills without task hindsight.

\subsection{Benchmark Construction Pipeline}

The construction of SkillGenBench follows the pipeline shown in Figure~\ref{fig:benchmark_construction}. We collect two classes of source materials: fixed-commit repository snapshots and long-form document bundles. For repository-grounded instances, we prioritize repositories in which key procedures are distributed across code, configurations, scripts, and environment conventions. For document-grounded instances, we prioritize long documents whose procedural constraints span multiple sections and cannot be recovered from a single passage.

\textbf{Stage 1: Knowledge Graph Construction.}
We construct knowledge graphs to support the subsequent stages. Each graph
abstracts the raw corpus into entity-relation triples, communities of related
procedural evidence, and context summaries covering input schemas, domain rules,
output formats, and validation criteria.

\textbf{Stage 2: Scenario Generation.}
From the knowledge graph and context summaries, we derive candidate scenarios
across several common task forms, such as code development, workflow execution,
and rule-grounded reasoning. Each scenario identifies a target workflow and the
relevant corpus evidence.

\textbf{Stage 3: Tasks and Test Cases Generation.}
Each scenario is used to generate a task specification and a set of test cases
covering normal, edge, and adversarial inputs. A self-reflection step then
refines each candidate for clarity and consistency.

\textbf{Stage 4: Task Verification without Skills.}
We discard tasks that are either solvable without procedural extraction or
trivially solvable. Specifically, we run two checks with a strong base model
(e.g., GPT-5): a \emph{corpus-free check}, where the model attempts the task
using only its parametric knowledge, and a \emph{with-corpus check}, where
the model is given the full source materials. Tasks with a pass rate $\geq 20\%$
on the corpus-free check or $\geq 50\%$ on the with-corpus check are returned
to Stage 3 for refinement.

\textbf{Stage 5: Task Verification with Skills.}
For each remaining task, we generate a reference skill and refine it through
iterative test-case feedback. We then run the task using this skill. If the
task fails to pass even with the reference skill, it is judged unrealistic
or overly hard, and is sent back to Stage 3.

This process repeats until the task falls within the target difficulty range or reaches the iteration limit. Accepted tasks then undergo a final human review (Appendix~\ref{app:human_verification}). Task candidates that do not pass validation are rewritten, refined, or replaced. The resulting instances are context-dependent, sufficiently challenging, and programmatically verifiable. They also share the same task format across heterogeneous repository and document sources, enabling direct comparison of skill-generation methods.

\subsection{Evaluation Protocol}

SkillGenBench evaluates a generated skill by its downstream behavior. An executor loads the skill and attempts the task. As summarized in Table~\ref{tab:benchmark_comparison}, instances fall into two evaluation modes, \emph{execution-based} and \emph{artifact-based}.

During skill generation, the model has access to the source materials and, in the task-conditioned setting, the task specification. Test cases, verifier internals, reference outputs, and held-out tasks are never exposed to the model. During execution, all generated skills are run in containerized environments under the same executor.

\paragraph{Execution-based evaluation.} The submitted code is run against hidden test cases with deterministic expected outputs, analogous to program-judging benchmarks. This mode is used when the desired result is a callable procedure or reusable implementation.

\paragraph{Artifact-based evaluation.} The submitted code is first executed to produce an artifact, which is then compared against a reference output. Comparison methods depend on the output modality, including exact matching, pixel-level similarity, semantic similarity, or an LLM judge when multiple valid outputs cannot be captured by a single deterministic metric. A heuristic pre-check (for example, resolution, duration, schema, or file format) filters out invalid outputs before comparison. This mode does not assume a unique ground-truth implementation, since many tasks admit multiple valid programs producing equivalent outputs.

\section{Experiments}
\label{sec:experiments}

\subsection{Experimental Setup}
\label{sec:exp_setup}

We evaluated five skill-generation baselines on SkillGenBench, selected to cover prompt-based generation, workflow-based generation and self-evolving generation. For each method, we vary the skill-generation backbone while keeping the downstream executor fixed. Specifically, all generated skills are evaluated by MiniMax-2.5~\cite{minimaxm25} under the same SkillGenBench evaluation harness, and task success is determined by the instance-specific verifier. The details of the baseline method are provided in Appendix~\ref{app:baselines}.

For each skill generation method, we instantiate the generator with six backbone models: Claude Sonnet 4.5~\cite{claudesonnet45}, GPT-5~\cite{gpt5}, Kimi K2.5~\cite{kimik25}, GLM-5~\cite{glm5}, MiniMax-M2.7~\cite{minimaxm27}, and Qwen3.6-Plus~\cite{qwen36plus}. All agentic interactions are executed through the same Claude Code runtime~\cite{claudecode}, with the backend model swapped through a unified API routing layer. This keeps the tool interface, filesystem access, and skill-packing procedure fixed across generation backbones, while varying only the model used to drive the generator.

During downstream evaluation, we report \textbf{pass@3} as the primary metric: each generated skill is evaluated with up to three independent trials, and an instance is counted as solved if any trial passes the instance-specific verifier. All reported dynamic results use a 1800-second per-instance budget over the 187-task benchmark. The static skill-structure analysis covers the same six-backbone generated-skill inventory.

The analysis proceeds from task success to artifact diagnosis. We first report dynamic execution results across methods and backbones, summarize the dominant source-level patterns, then inspect the generated skill artifacts themselves through static diagnostics. Finally, we analyze completed verifier failures to explain which source-specific mechanisms remain unresolved.

\subsection{Dynamic Execution Results}
\label{sec:dynamic_results}

\begin{table}[t]
  \caption{Main pass@3 results (\%) split by source family. For each generation backbone, Code denotes Code Repo tasks and Doc combines Code Doc and Domain Knowledge Doc tasks. Avg. averages over the six backbones.}
  \label{tab:main_results}
  \centering
  \scriptsize
  \setlength{\tabcolsep}{2pt}
  \renewcommand{\arraystretch}{1.08}
  \resizebox{\linewidth}{!}{%
  \begin{tabular}{@{}l*{14}{c}@{}}
    \toprule
    \textbf{Method}
      & \multicolumn{2}{c}{\shortstack{\textbf{Sonnet}\\\textbf{4.5}}}
      & \multicolumn{2}{c}{\textbf{GPT-5}}
      & \multicolumn{2}{c}{\shortstack{\textbf{Kimi}\\\textbf{K2.5}}}
      & \multicolumn{2}{c}{\textbf{GLM-5}}
      & \multicolumn{2}{c}{\shortstack{\textbf{MiniMax}\\\textbf{M2.7}}}
      & \multicolumn{2}{c}{\shortstack{\textbf{Qwen3.6}\\\textbf{Plus}}}
      & \multicolumn{2}{c}{\textbf{Avg.}} \\
    \cmidrule(lr){2-3}\cmidrule(lr){4-5}\cmidrule(lr){6-7}\cmidrule(lr){8-9}\cmidrule(lr){10-11}\cmidrule(lr){12-13}\cmidrule(l){14-15}
      & \textbf{Code} & \textbf{Doc}
      & \textbf{Code} & \textbf{Doc}
      & \textbf{Code} & \textbf{Doc}
      & \textbf{Code} & \textbf{Doc}
      & \textbf{Code} & \textbf{Doc}
      & \textbf{Code} & \textbf{Doc}
      & \textbf{Code} & \textbf{Doc} \\
    \midrule
        \textsc{No Skill} & 13.8 & 23.4 & 13.8 & 23.4 & \textbf{13.8} & \textbf{23.4} & 13.8 & 23.4 & \textbf{13.8} & 23.4 & 13.8 & 23.4 & 13.8 & 23.4 \\
    \midrule
    \textsc{Naive Prompt} & 7.3 & 23.4 & 12.2 & 21.9 & 9.8 & 17.2 & \textbf{16.3} & 23.4 & 13.0 & 20.3 & 14.6 & 25.0 & 12.2 & 21.9 \\
    \textsc{EvoSkill} & 9.8 & \textbf{26.6} & 14.6 & \textbf{31.2} & 7.3 & 15.6 & 11.4 & 17.2 & 6.5 & 20.3 & 15.4 & \textbf{28.1} & 10.8 & 23.2 \\
    \textsc{SkillNet} & 11.4 & 23.4 & \textbf{17.9} & 21.9 & 9.8 & 18.8 & 13.8 & 21.9 & 5.7 & 14.1 & \textbf{16.3} & \textbf{28.1} & 12.5 & 21.4 \\
    \textsc{SkillCreator} & 10.6 & 25.0 & 14.6 & 23.4 & 8.9 & 20.3 & \textbf{16.3} & 26.6 & 7.3 & 18.8 & 15.4 & 23.4 & 12.2 & 22.9 \\
    \textsc{SkillSeekers} & \textbf{16.3} & 25.0 & 17.1 & 28.1 & 9.8 & \textbf{23.4} & 14.6 & \textbf{28.1} & 13.0 & \textbf{21.9} & 15.4 & 23.4 & \textbf{14.4} & \textbf{25.0} \\
    \bottomrule
  \end{tabular}
  }
\vspace{-0.2in}
\end{table}

Table~\ref{tab:main_results} summarizes the main dynamic results. Across the six generation backbones, \textsc{SkillSeekers} achieves the best average performance (14.4\% on Code and 25.0\% on Doc). In several Code settings, prompt-only generation remains competitive; moreover, when the LLM backbone is relatively weak, even more sophisticated pipelines such as \textsc{SkillCreator} struggle to achieve strong performance. These results indicate that improvements from skill-generation methods are not stable, and depend critically on the interaction between the generator, backbone model, and source type.

More importantly, under strict execution-based evaluation, generated skills are not universally beneficial and can in some cases perform worse than no-skill baselines. This typically occurs when the generated artifact introduces interface inconsistencies, incomplete procedures, or incorrect assumptions that interfere with the executor’s parametric knowledge. In contrast, skills are most helpful when they provide precise, source-grounded procedures that the base model cannot easily infer.

A consistent pattern across all methods is the substantial gap between Code and Doc tasks. Code performance remains low (10.8\%–14.4\%), while Doc performance is significantly higher (21.4\%–25.0\%). This reflects the additional challenge of repository-grounded skill generation, where models must recover implicit execution structure—such as environment setup, command conventions, and data flow—from distributed code artifacts.

\begin{wrapfigure}{r}{0.48\linewidth}
  \centering
  \vspace{-10pt}
  \includegraphics[width=\linewidth]{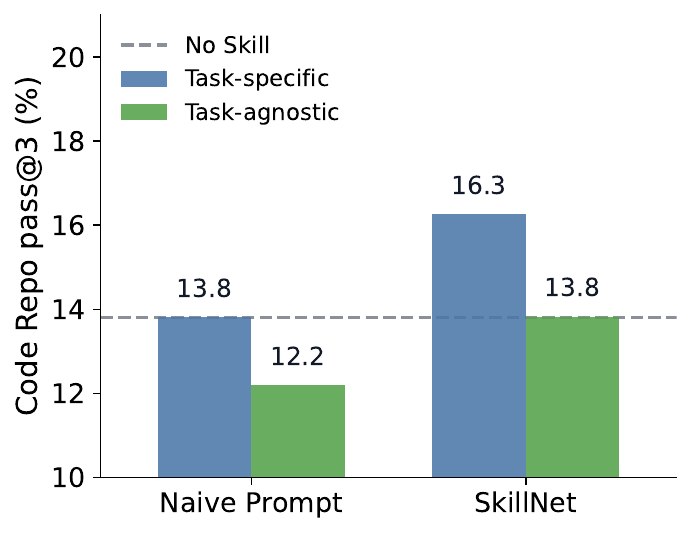}
  \caption{Repository-grounded task-specific versus task-agnostic pass@3 results for GLM-5 and Qwen3.6-Plus. Bars compare generation regimes within the same method and backbone.}
  \label{fig:regime_comparison_repo}
  \vspace{-10pt}
\end{wrapfigure}

Figure~\ref{fig:regime_comparison_repo} further highlights the limitations of task-agnostic skill generation. Without task-specific guidance, generators must distill broadly reusable procedural knowledge, which remains challenging for current methods and moderately capable backbones. As a result, task-agnostic skills often fail to capture the precise constraints required for downstream execution, leading not only to weaker performance than task-conditioned generation, but in some cases even underperforming the no-skill baseline. This suggests that unconstrained skill abstraction may produce artifacts that are structurally plausible but poorly aligned with actual execution requirements, resulting in negative transfer.

Appendix Figure~\ref{fig:token_limit_sensitivity} shows that increasing the generation budget improves performance up to a point (roughly 24K–64K tokens), after which gains saturate, indicating that generation capacity alone is insufficient to overcome these limitations.

\subsection{Static Analysis}
\label{sec:static_structure}

Dynamic pass@3 measures whether a generated skill helps the fixed executor solve a task, but it does not reveal what kind of artifact each generator produces. We therefore supplement execution results with static diagnostics over the generated skill packages.

\begin{figure}[!htbp]
  \centering
  \begin{minipage}[t]{0.48\textwidth}
    \centering
    \includegraphics[width=0.95\linewidth, height=2.2in, keepaspectratio]{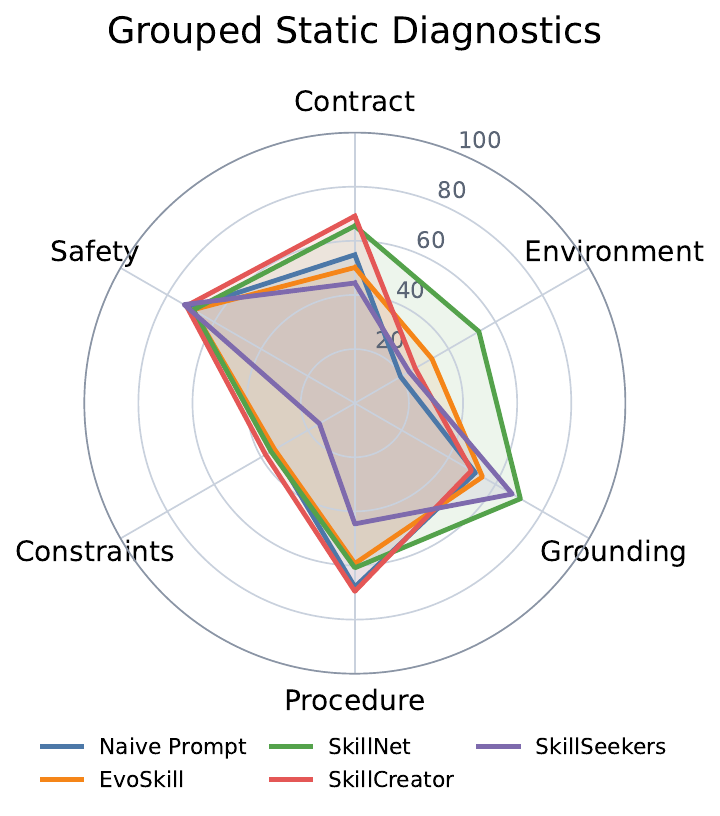}
    \caption{Grouped static diagnostics over generated skill packages. Axes aggregate automatic rule-based checks; higher is better.}
    \label{fig:static_radar}
  \end{minipage}
   \hfill
  \begin{minipage}[t]{0.5\textwidth}
    \centering
    \includegraphics[width=0.95\linewidth, height=2.2in, keepaspectratio]{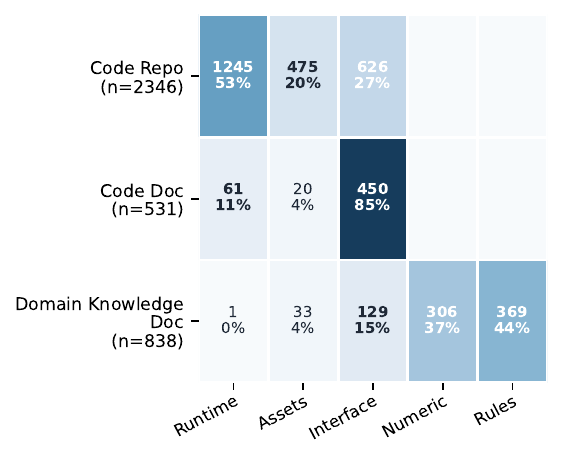}
    \caption{Completed verifier-failure taxonomy. Cells give counts and shares; row labels give totals.}
    \label{fig:completed_failure_taxonomy}
  \end{minipage}
  \vspace{-0.1in}
\end{figure}

Each skill is first scored by eight automatic rule-based diagnostics over the generated \texttt{SKILL.md} package and grouped into six main-paper axes. \emph{Contract} averages interface-contract and verification cues. \emph{Environment} measures setup and dependency readiness. \emph{Grounding} measures explicit ties to source artifacts. \emph{Procedure} averages procedural coverage and state/data handling. \emph{Constraints} measures whether strict task rules are preserved. \emph{Safety} measures artifact hygiene, including conciseness and avoidance of brittle task-specific leakage or risky commands.

\begin{table}
  \caption{Grouped static skill scores by method. Scores are averaged over the canonical six-backbone generated-skill inventory and reported on a 0--100 scale. Overall averages the six grouped diagnostics.}
  \label{tab:static_scores}
  \centering
  \setlength{\tabcolsep}{2.4pt}
  \renewcommand{\arraystretch}{1.12}
{
  \begin{tabular}{lrrrrrrrr}
    \toprule
    \textbf{Method}
      & \textbf{N}
      & \textbf{Overall}
      & \textbf{Contract}
      & \textbf{Env.}
      & \textbf{Ground}
      & \textbf{Proc.}
      & \textbf{Const.}
      & \textbf{Safety} \\
    \midrule
    \textsc{Naive Prompt} & 1129 & 49.8 & 54.8 & 19.5 & 51.5 & 67.9 & 34.6 & 70.3 \\
    \textsc{EvoSkill} & 1093 & 50.0 & 50.1 & 32.9 & 54.5 & 59.3 & 34.2 & 69.0 \\
    \textsc{SkillNet} & 1117 & \textbf{59.1} & 65.5 & \textbf{52.9} & \textbf{70.7} & 60.9 & 35.8 & 68.9 \\
    \textsc{SkillCreator} & 1109 & 54.0 & \textbf{69.1} & 25.8 & 49.8 & \textbf{69.5} & \textbf{38.1} & 71.9 \\
    \textsc{SkillSeekers} & 1138 & 44.6 & 44.4 & 23.2 & 67.2 & 44.6 & 15.1 & \textbf{72.8} \\
    \bottomrule
  \end{tabular}
  }
\vspace{-0.1in}
\end{table}

Figure~\ref{fig:static_radar} and Table~\ref{tab:static_scores} show qualitatively different artifacts. \textsc{SkillNet} has the strongest grouped static score, driven by Environment and Grounding. \textsc{SkillCreator} is strongest on Contract, Procedure, and Constraints. \textsc{SkillSeekers}, despite the strongest Code and Doc averages in Table~\ref{tab:main_results}, has the best Safety score and strong Grounding but weaker Contract, Procedure, and Constraints.

This mismatch indicates that static quality and execution success capture different aspects of skill quality: the former focuses on structural completeness, while the latter tests whether the skill can be executed correctly. Therefore, structural completeness does not guarantee executability, and dynamic success does not necessarily imply sound structure. This finding further suggests that the core challenge in skill generation lies in bridging the gap between specification and execution; optimizing either side alone is insufficient.

\subsection{Error Analysis}
\label{sec:completed_failure_analysis}

Aggregate pass rates do not explain why a generated skill still fails after it is invoked. We therefore inspect completed verifier failures—cases where the executor produces a concrete answer under a generated skill, but the instance verifier rejects the output. This isolates errors in the distilled procedure and its operationalization, rather than counting cases where the skill is never exercised.

Figure~\ref{fig:completed_failure_taxonomy} applies a source-aware failure taxonomy across the three procedural sources for the 1800-second evaluation batch. Code Repo failures are dominated by runtime or dependency issues (1245, 53\%), followed by interface or schema errors (626, 27\%) and asset or artifact issues (475, 20\%). Code Doc failures are more concentrated: interface or schema errors account for 450 failures (85\%), with a smaller runtime or dependency bucket (61, 11\%). Domain Knowledge Doc failures exhibit a different profile, with state or rule errors (369, 44\%) and numeric or formula errors (306, 37\%) dominating, and fewer interface or schema errors (129, 15\%).

This taxonomy clarifies why dynamic and static results diverge. Code Repo failures are largely driven by execution environment, asset, and interface issues, so improved textual grounding alone does not guarantee success. Code Doc failures mostly reduce to schema and format precision, where explicit interface contracts and verification cues are critical. Domain Knowledge Doc failures instead require precise numeric, state, and rule encoding, which is only weakly captured by coarse procedural coverage.

Overall, the results support four conclusions. First, skill generation should be evaluated as a generator--backbone--executor pipeline rather than as an isolated prompting recipe. Second, the main difficulty is source-specific: repository tasks require operational recovery, code documentation tasks require exact interface compliance, and domain documents require precise rule execution. Third, task-agnostic skills can help, but only when they preserve transferable procedures without discarding task-specific constraints. Fourth, artifact diagnostics are necessary for explanation, but execution-based pass@3 remains the decisive measure of whether a generated skill is actually useful.


\section{Conclusion}
\label{sec:conclusion}

We introduced SkillGenBench, a benchmark for evaluating skill generation as a first-class problem in LLM agent systems. By decoupling upstream skill generation from downstream execution, SkillGenBench enables controlled comparison of procedure-to-skill distillation pipelines across repository and document sources.

Our experiments show that skill generation is fundamentally a pipeline-level problem: performance depends not only on the generation method, but also on the backbone model and the nature of the source material. In particular, repository-grounded tasks remain significantly more challenging than document-based ones, highlighting the difficulty of recovering implicit execution structure from distributed code artifacts.
More importantly, we identify a persistent gap between specification and execution. Generated skills often capture the right structural components, yet fail to translate them into executable procedures that satisfy strict verification constraints. This gap is especially pronounced in settings that require precise interface alignment, state handling, and rule fidelity.

These findings suggest that improving skill generation requires going beyond surface-level structure and addressing execution-level correctness. Static diagnostics and execution-based evaluation therefore play complementary roles: the former explains what a skill contains, while the latter determines whether it actually works.

SkillGenBench provides both a benchmark and an analysis framework for studying this gap. We hope it will encourage future work to focus not only on generating skills, but on ensuring that they are executable, reliable, and aligned with real-world procedural constraints.

\bibliography{references}


\appendix



\newpage

\section{Additional Results}

\subsection{Method--Backbone Heatmap}
\label{app:pass3_heatmap}

\begin{figure}[h]
  \centering
  \includegraphics[width=0.8\linewidth]{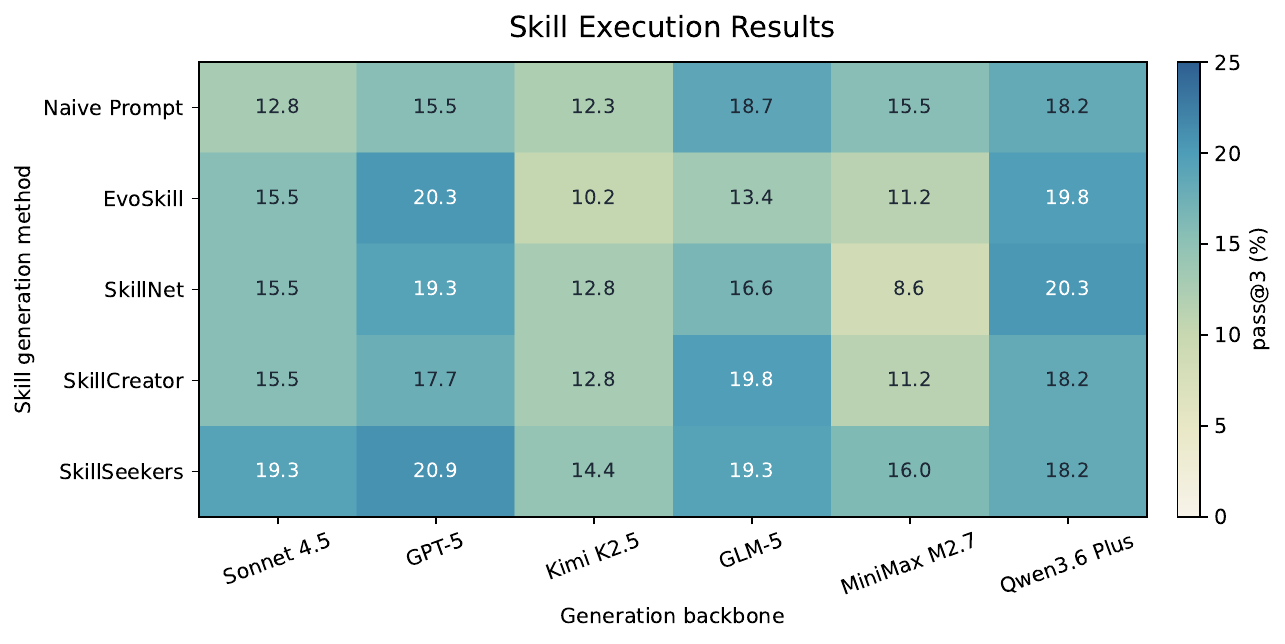}
  \caption{Full method--backbone pass@3 matrix across skill-generation methods and generation backbones. The downstream executor is fixed; only the upstream skill generator and its backbone vary.}
  \label{fig:pass3_heatmap}
\end{figure}

Figure~\ref{fig:pass3_heatmap} reports the full pass@3 matrix across the five skill-generation methods and the six generation backbones, complementing the aggregate Code/Doc columns of Table~\ref{tab:main_results}. The downstream executor is held fixed; only the skill generator and its backbone vary.

Two patterns are worth noting. First, no method dominates uniformly:
\textsc{SkillSeekers} attains the highest pass@3 on Sonnet 4.5, GPT-5,
Kimi K2.5, and MiniMax M2.7, but \textsc{SkillNet} leads on Qwen3.6 Plus (20.3\%) and \textsc{SkillCreator} is on par with \textsc{SkillSeekers} on GLM-5 (19.8\% vs.\ 19.3\%). Second, methods differ in their across-backbone spread: \textsc{SkillSeekers} stays within a 14.4--20.9\% band, while \textsc{EvoSkill} ranges from 10.2\% (Kimi K2.5) to 20.3\% (GPT-5), indicating that some pipelines are more sensitive to the choice of backbone than others.

\begin{figure}[h]
  \centering
  \includegraphics[width=0.8\linewidth]{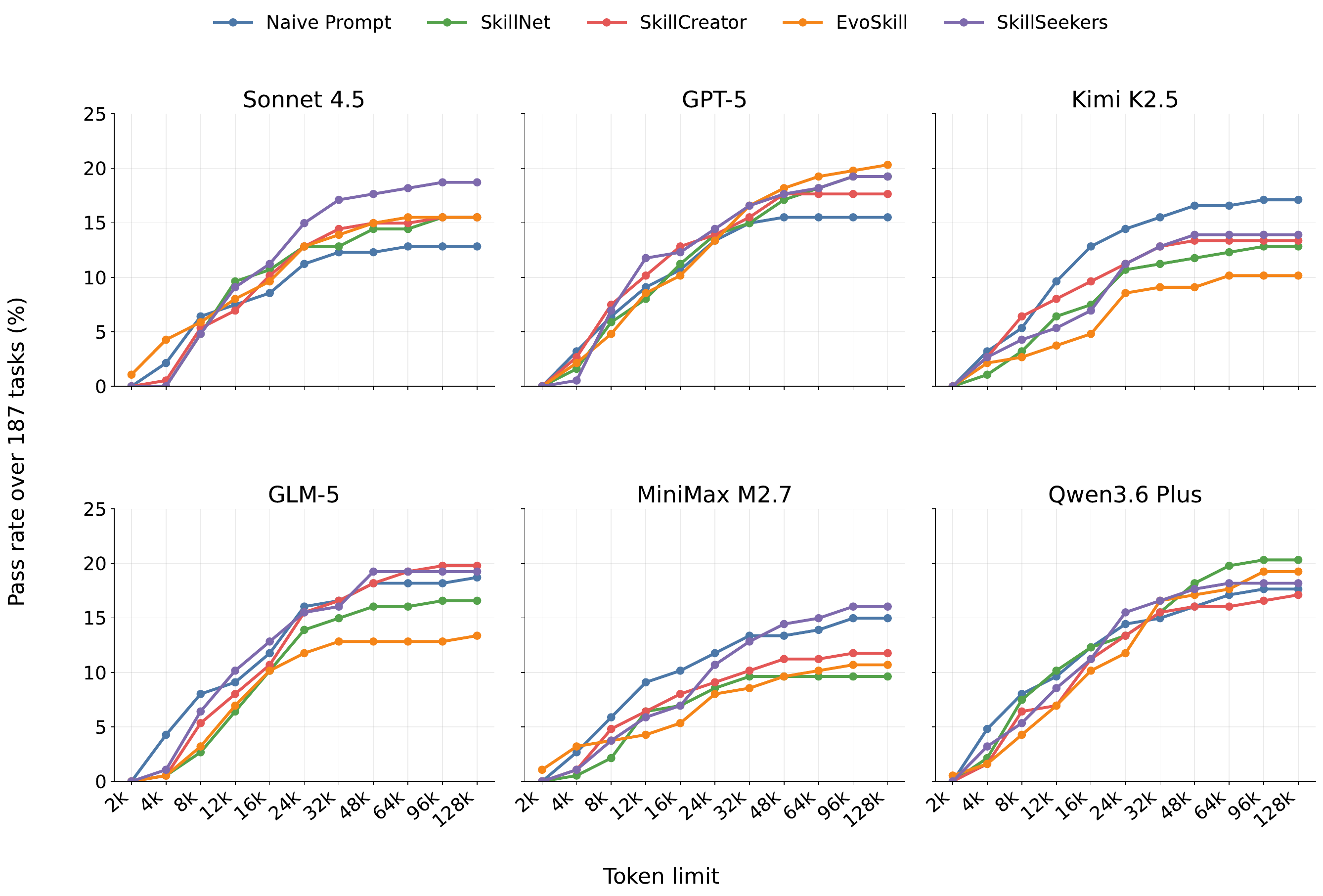}
  \caption{Sensitivity of benchmark pass rate to the generation token limit. Each panel fixes the generation backbone and plots pass rate over the same 187-task suite as the available token budget increases.}
  \label{fig:token_limit_sensitivity}
\end{figure}

\subsection{Token-Limit Sensitivity}
\label{app:token_limit_sensitivity}

Figure~\ref{fig:token_limit_sensitivity} examines how the generation token budget influences pass@3 on the full 187-task suite. For each backbone, we sweep token budgets from 2K to 128K and report the pass@3 achievable within each budget across all five skill-generation methods, while keeping the executor and all other settings fixed.

Across backbones, the pass rate rises steeply up to roughly 16K--24K tokens, then flattens between 32K and 64K, with little additional gain at 96K or 128K. The plateau height, however, is backbone-dependent: GPT-5 and GLM-5 saturate near 18--20\% pass@3, whereas Kimi K2.5 and MiniMax M2.7 plateau between 10\% and 17\%. This plateau reinforces the observation in Section~\ref{sec:experiments} that additional generation budget alone does not close the remaining gap.

\subsection{Bootstrap Confidence Intervals}
\label{app:bootstrap_ci}

Table~\ref{tab:bootstrap_ci} reports task-level bootstrap 95\% confidence intervals for the overall pass@3 results in Table~\ref{tab:main_results}. We draw $B{=}2000$ bootstrap resamples (with replacement) over the 187 benchmark tasks, computing pass@3 on each resample. The CI half-width is approximately $\pm$5 percentage points across all cells, indicating that most pairwise method differences are not statistically distinguishable at this benchmark scale. This supports the conclusion that skill-generation method choice and backbone choice together drive performance, and that no single method dominates across all backbones.

\begin{table}[h]
  \caption{Bootstrap 95\% confidence intervals for overall pass@3 (\%). Each cell shows mean [CI$_\mathrm{lo}$, CI$_\mathrm{hi}$] estimated from $B{=}2000$ task-level bootstrap resamples over 187 tasks.}
  \label{tab:bootstrap_ci}
  \centering
  \scriptsize
  \setlength{\tabcolsep}{3pt}
  \renewcommand{\arraystretch}{1.12}
  \resizebox{\linewidth}{!}{%
  \begin{tabular}{lcccccc}
    \toprule
    \textbf{Method} & \textbf{Sonnet 4.5} & \textbf{GPT-5} & \textbf{Kimi K2.5} & \textbf{GLM-5} & \textbf{MiniMax M2.7} & \textbf{Qwen3.6 Plus} \\
    \midrule
    \textsc{Naive Prompt} & 12.8 [8.0, 17.6] & 15.5 [10.7, 20.9] & 17.1 [11.8, 22.5] & 18.7 [13.4, 24.6] & 15.5 [10.7, 21.4] & 17.6 [12.3, 23.5] \\
    \textsc{SkillNet} & 15.5 [10.2, 21.4] & 19.2 [13.4, 24.6] & 12.8 [8.0, 17.7] & 16.6 [11.8, 21.9] & 10.2 [5.9, 15.0] & 20.3 [14.4, 26.2] \\
    \textsc{SkillCreator} & 15.5 [10.7, 20.9] & 17.6 [12.3, 23.5] & 13.4 [9.1, 18.7] & 19.8 [14.4, 25.7] & 11.8 [7.5, 16.6] & 17.6 [12.3, 23.0] \\
    \textsc{EvoSkill} & 15.5 [10.7, 20.9] & 20.3 [15.0, 26.2] & 10.2 [5.9, 14.4] & 13.4 [9.1, 18.2] & 11.2 [7.0, 15.5] & 19.2 [13.9, 25.1] \\
    \textsc{SkillSeekers} & 19.2 [13.9, 24.6] & 19.2 [13.9, 25.1] & 14.4 [9.6, 19.8] & 19.2 [13.9, 25.1] & 16.0 [10.7, 21.4] & 18.2 [12.8, 23.5] \\
    \bottomrule
  \end{tabular}
  }
\end{table}

\section{Model and Harness Configurations}
\label{app:runtime_details}

\paragraph{Runtime harness.}
All experiments use the same SkillGenBench harness and Claude Code runtime. Skill generation is instantiated with the six backbone models described in Section~\ref{sec:exp_setup}, while downstream execution is fixed to MiniMax-2.5. We run agent interactions through Claude Code CLI 2.1.85 with \texttt{claude-agent-sdk} 0.1.64.

\paragraph{Execution environment.}
Downstream evaluations are executed in isolated Docker environments selected by the instance configuration. No GPU resources are requested.

\paragraph{Skill-generation stage model hyperparameters.}
\begin{itemize}
    \item Temperature: 0
    \item Max output tokens: 16,384
    \item Max rounds: 3 refinement iterations / 45 agent turns
    \item Timeout: 1800 seconds
\end{itemize}

\paragraph{Evaluation stage model hyperparameters.}
All downstream evaluations use the same executor-side generation setting with MiniMax-2.5:
\begin{itemize}
    \item Temperature: 0
    \item Max output tokens: 16,384
    \item Timeout: 1800 seconds
\end{itemize}

\section{Baseline Methods}
\label{app:baselines}

We compare five skill-generation baselines that cover the main ways current systems construct reusable agent skills. Following the experimental setup in Section~\ref{sec:exp_setup}, we organize them into three families: prompt-based generation, workflow-based generation, and self-evolving generation. All baselines operate under the same SkillGenBench visibility boundary and output the same \texttt{SKILL.md} skill package, so the comparison focuses on the skill-construction procedure rather than differences in downstream execution.

\paragraph{Prompt-based skill generation.} \textsc{Naive Prompt} is the minimal prompt-based baseline. The generator receives the visible corpus and, in the task-conditioned setting, the task instruction, then directly writes a skill package in a single generation pass. This baseline measures how much reusable procedural knowledge can be distilled from the exposed materials without trajectories, search, self-evaluation, or an explicit skill-authoring workflow.

\paragraph{Workflow-based skill generation.} We include three system-level baselines that construct skills through explicit workflows. \textsc{SkillNet}~\cite{skillnet} represents toolkit-mediated skill creation, where source materials are transformed through a dedicated skill creation interface. \textsc{SkillSeekers}~\cite{skillseekers} represents source-to-skill conversion pipelines for repositories and documents, emphasizing the extraction and packaging of actionable knowledge from external materials. \textsc{SkillCreator}~\cite{skillcreator} represents iterative skill authoring, where an agent drafts, evaluates, and refines a skill before submission. These baselines share the same final interface, but differ in the inductive bias imposed by the construction process: toolkit-based packaging, source conversion, and self-refined authoring.

\paragraph{Self-evolving skill generation.} \textsc{EvoSkill}~\cite{evoskill} represents methods that derive skills from execution experience rather than static context alone. In SkillGenBench, EvoSkill receives the visible corpus together with trajectories collected from corresponding runs without generated skills. This setting tests whether observed execution behavior provides useful procedural evidence for skill generation while preserving the benchmark boundary: trajectories are generated from the same visible task environment, and hidden tests or verifier internals are never exposed.

\paragraph{Unified adaptation.} For released systems, we follow their official workflows and recommended settings whenever they are applicable to the SkillGenBench interface. Adaptations are restricted to benchmark integration: formatting the visible input bundle for each method, routing model calls through the shared backend, and normalizing outputs into the standardized \texttt{SKILL.md} layout. In the task-conditioned setting, the generator receives task-specific materials; in the task-agnostic setting, it receives only collection-level materials and must produce a reusable skill before downstream tasks are revealed. Our EvoSkill instantiation uses the vendored proposer--generator assets with benchmark-collected trajectories; it should be interpreted as a SkillGenBench adaptation of EvoSkill rather than a full reproduction of its native multi-round self-improving loop.

\section{Case Studies}
\label{app:case_studies}

We present representative benchmark items from SkillGenBench to illustrate how procedural knowledge encoded in skills affects downstream task execution.

\begin{tcolorbox}[
    breakable,
    title={Case 1: StyleTransfer\_gtb01 --- Neural Style Transfer (Image)},
    colback=LighterGray,
    colframe=DeepPurple,
    colbacktitle=DeepPurple,
    coltitle=white,
]
\textbf{\emph{\textcolor{DeepPurple}{Task}}}

Apply the style of a given artwork to a content photograph using the StyleTransfer repository, and save the stylized output as \texttt{styled\_image.jpg}.

\textbf{\emph{\textcolor{DeepPurple}{Skill-Dependent Knowledge}}}

The generated skill must capture the repo's command-line interface, default optimization parameters, and output naming conventions. Without the skill, a generic agent may invoke the wrong entry point, use incorrect argument names, or save the output to an unexpected location.

\textbf{\emph{\textcolor{DeepPurple}{Inputs and Reference Output}}}

\begin{center}
\begin{minipage}[b]{0.30\linewidth}
  \centering
  \includegraphics[width=\linewidth]{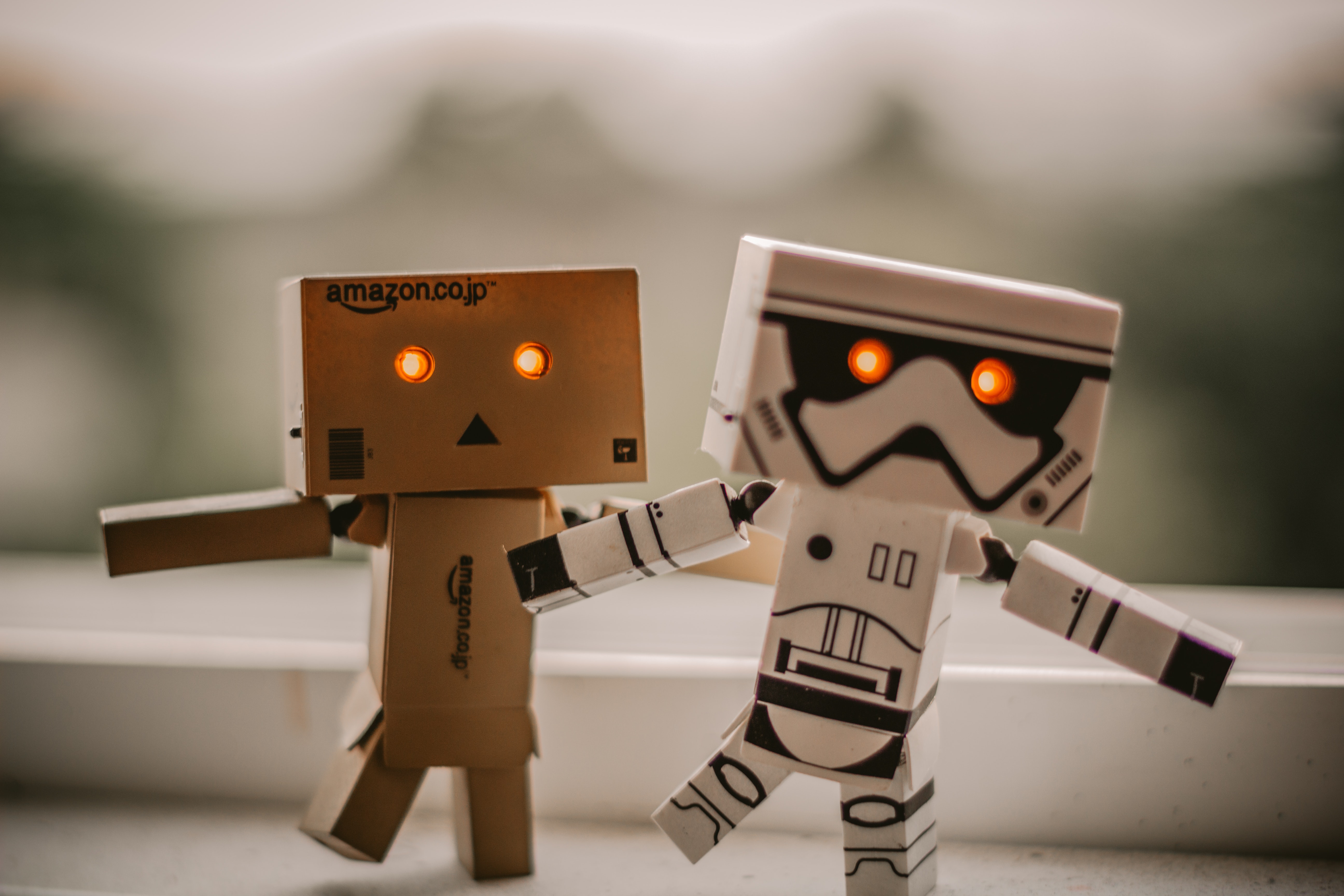}\\[2pt]
  {\small (a) Content image}
\end{minipage}
\hfill
\begin{minipage}[b]{0.30\linewidth}
  \centering
  \includegraphics[width=\linewidth]{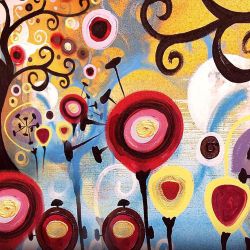}\\[2pt]
  {\small (b) Style image}
\end{minipage}
\hfill
\begin{minipage}[b]{0.30\linewidth}
  \centering
  \includegraphics[width=\linewidth]{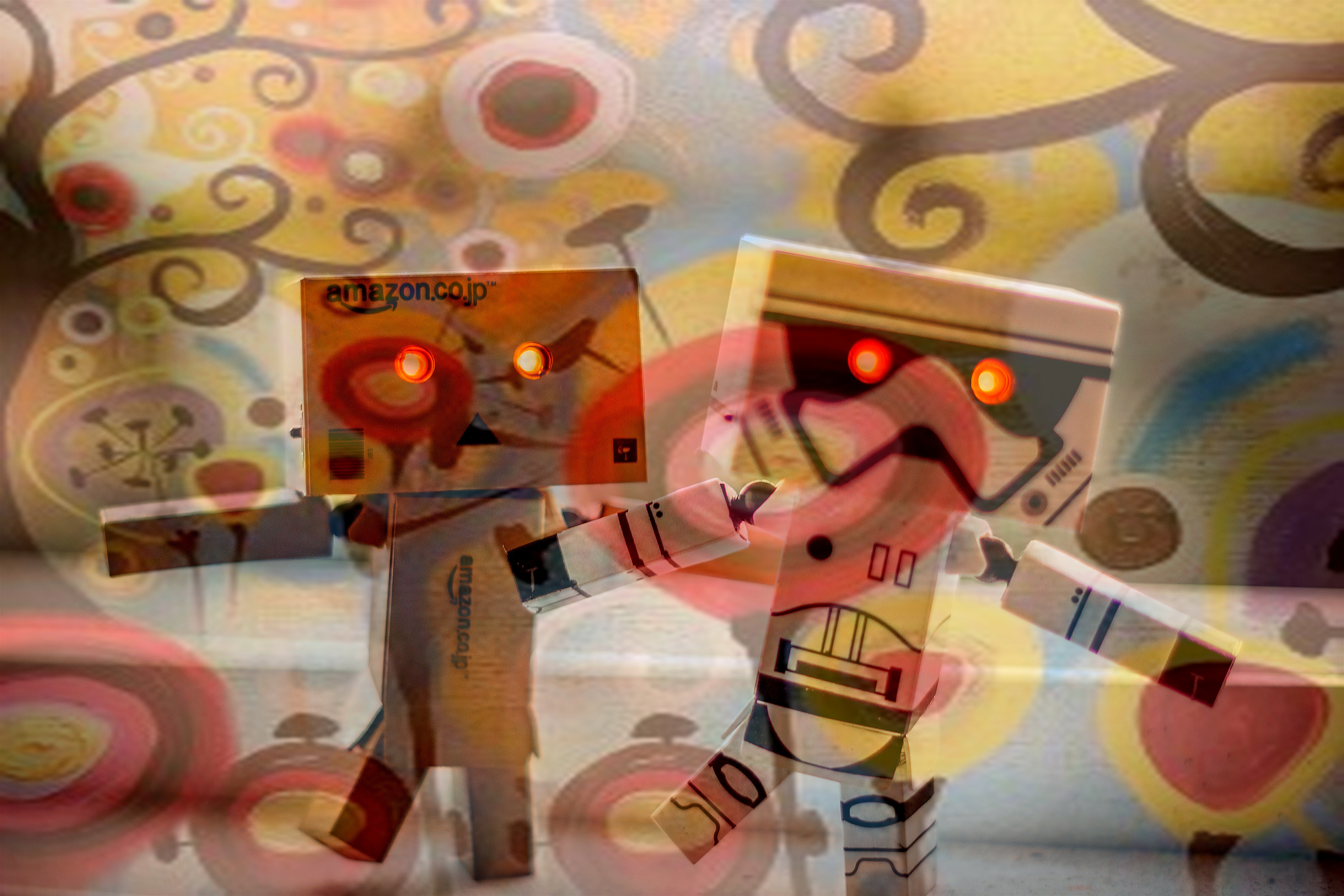}\\[2pt]
  {\small (c) Stylized output}
\end{minipage}
\end{center}

The stylized result transfers the vibrant swirling patterns and warm color palette of the style painting onto the content photograph, while preserving the overall structure and recognizable shapes of the robot figurines.
\end{tcolorbox}

\begin{tcolorbox}[
    breakable,
    title={Case 2: AnimeGANv3\_gen04 --- Anime Style Transformation (Image)},
    colback=LighterGray,
    colframe=DeepPurple,
    colbacktitle=DeepPurple,
    coltitle=white,
]
\textbf{\emph{\textcolor{DeepPurple}{Task}}}

Transform a mountain landscape photograph into Shinkai Makoto animation style using the AnimeGANv3 library, preserving the original $2048{\times}1365$ pixel dimensions. Save the result as a properly color-corrected PNG file.

\textbf{\emph{\textcolor{DeepPurple}{Skill-Dependent Knowledge}}}

AnimeGANv3 uses BGR color channel ordering internally, diverging from the standard RGB convention. The repo provides explicit color space conversion utilities that must be applied during pre- and post-processing. An agent unfamiliar with this requirement will produce images with swapped color channels (warm tones appear cool and vice versa). Additionally, the agent must select the correct Shinkai-specific ONNX model variant from among multiple style options.

\textbf{\emph{\textcolor{DeepPurple}{Inputs and Reference Output}}}

\begin{center}
\begin{minipage}[b]{0.45\linewidth}
  \centering
  \includegraphics[width=\linewidth]{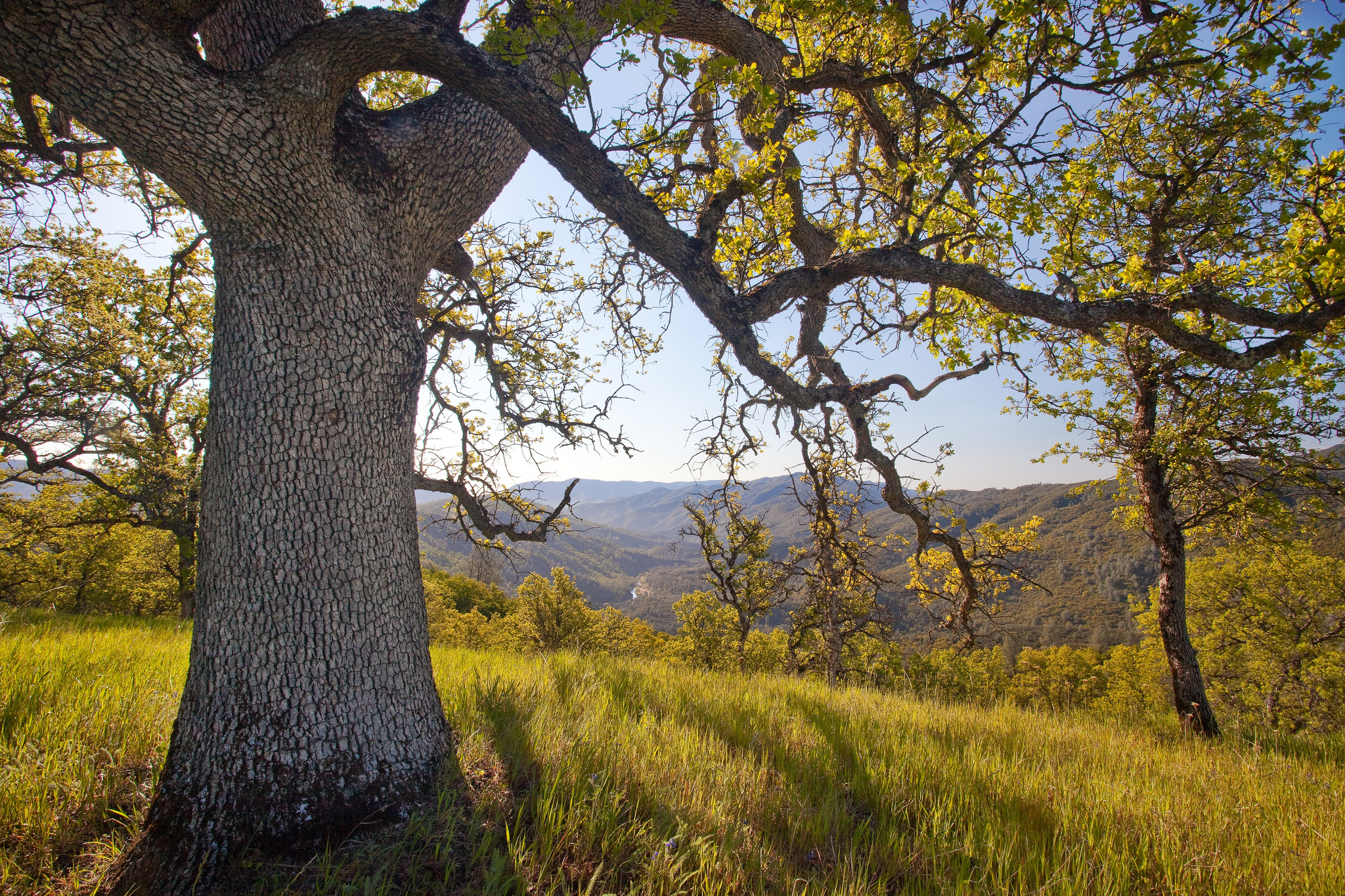}\\[2pt]
  {\small (a) Input photograph}
\end{minipage}
\hfill
\begin{minipage}[b]{0.45\linewidth}
  \centering
  \includegraphics[width=\linewidth]{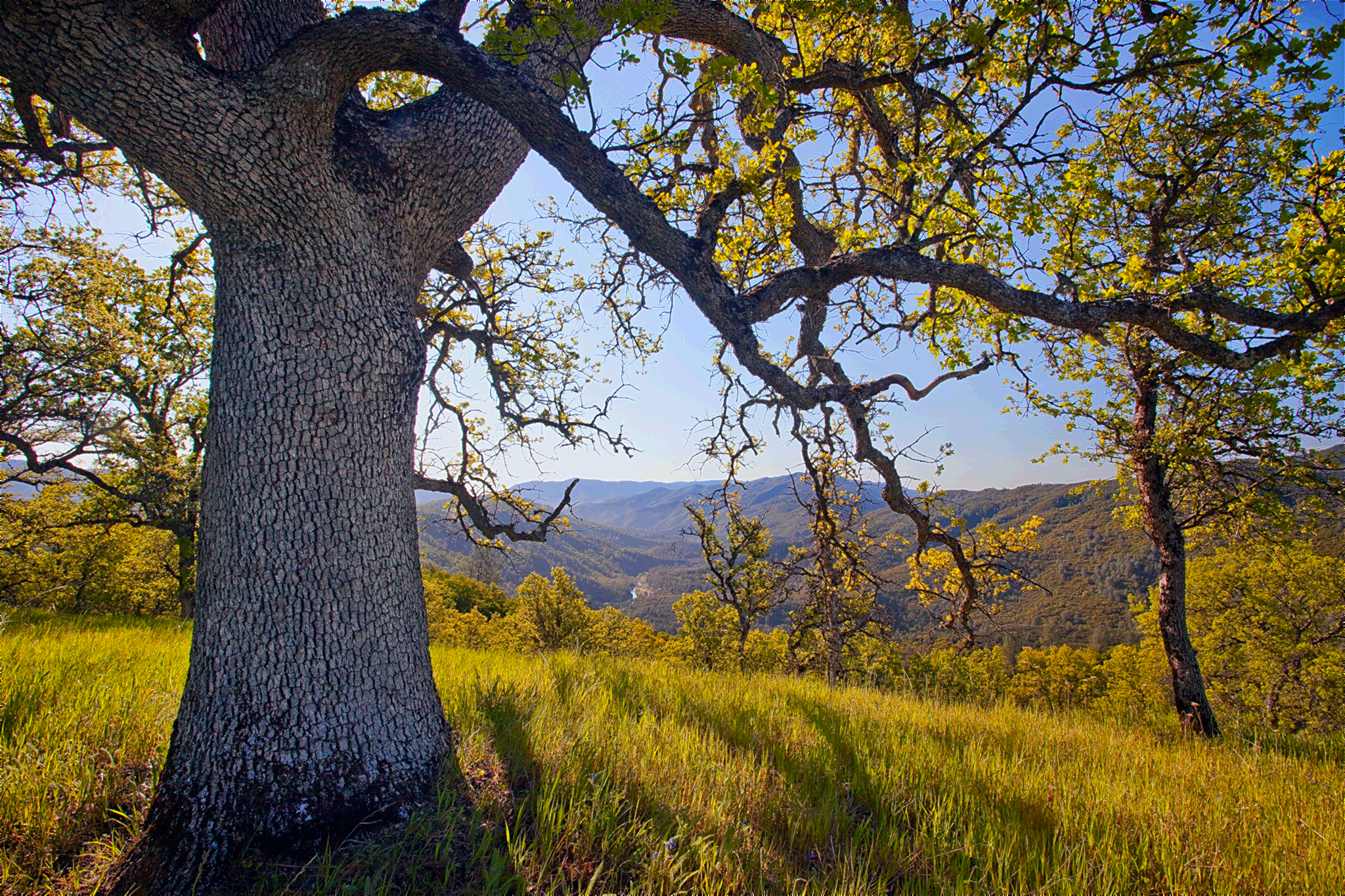}\\[2pt]
  {\small (b) Shinkai-style output}
\end{minipage}
\end{center}

The output exhibits characteristic Shinkai-style atmospheric effects, enhanced color grading, and stylized lighting while maintaining the original image dimensions.
\end{tcolorbox}

\begin{tcolorbox}[
    breakable,
    title={Case 3: Faker\_gen08 --- Synthetic User Profile Generation (Code)},
    colback=LighterGray,
    colframe=DeepPurple,
    colbacktitle=DeepPurple,
    coltitle=white,
]
\textbf{\emph{\textcolor{DeepPurple}{Task}}}

Using the Faker library, generate a JSON file containing exactly 100 synthetic user profiles for an e-commerce platform. Each profile must include \texttt{username}, \texttt{email}, \texttt{ipv4}, and \texttt{user\_agent} fields. All usernames must be unique. The output must be deterministically reproducible under \texttt{Faker.seed(24680)}.

\textbf{\emph{\textcolor{DeepPurple}{Skill-Dependent Knowledge}}}

\textbf{\textcolor{CaseOrange}{1.}} Faker's \texttt{fake.unique.user\_name()} proxy must be used instead of \texttt{fake.user\_name()} to guarantee username uniqueness---without it, duplicates may silently appear.

\textbf{\textcolor{CaseOrange}{2.}} The correct snake\_case method names (\texttt{user\_name}, \texttt{ipv4}) differ from common alternatives (\texttt{userName}, \texttt{ipv4\_address}) that would raise errors or produce different outputs.

\textbf{\textcolor{CaseOrange}{3.}} Class-level seeding via \texttt{Faker.seed(24680)} must be called before any generation to ensure byte-exact reproducibility against the reference output.

\textbf{\textcolor{CaseOrange}{4.}} Serialization must use \texttt{json.dumps(..., indent=2, ensure\_ascii=False)} to match the reference format.

\textbf{\emph{\textcolor{DeepPurple}{Expected Output (excerpt)}}}

{\small
\begin{verbatim}
[
  {
    "username": "sarah34",
    "email": "michael45@example.com",
    "ipv4": "192.168.1.42",
    "user_agent": "Mozilla/5.0 ..."
  },
  ...  // 100 objects total, all usernames unique
]
\end{verbatim}
}
\end{tcolorbox}

\begin{tcolorbox}[
    breakable,
    title={Case 4: PDFPlumber\_gen03 --- PDF Text Extraction Statistics (Code)},
    colback=LighterGray,
    colframe=DeepPurple,
    colbacktitle=DeepPurple,
    coltitle=white,
]
\textbf{\emph{\textcolor{DeepPurple}{Task}}}

Using the PDFPlumber library, analyze a technical manual PDF and produce a JSON file listing, for each page, the \texttt{page\_number}, \texttt{word\_count}, and \texttt{line\_count}.

\textbf{\emph{\textcolor{DeepPurple}{Skill-Dependent Knowledge}}}

\textbf{\textcolor{CaseOrange}{1.}} Word counting must use \texttt{page.extract\_words()}, which returns word-level bounding box objects---not naive whitespace splitting of extracted text, which produces different counts around punctuation and ligatures.

\textbf{\textcolor{CaseOrange}{2.}} Line counting must use \texttt{page.extract\_text()} with the default \texttt{layout=False} setting and split on newline characters. Using \texttt{layout=True} or inferring lines from word y-coordinates yields different counts.

\textbf{\textcolor{CaseOrange}{3.}} The JSON output must preserve field order (\texttt{page\_number}, \texttt{word\_count}, \texttt{line\_count}) and page ordering.

\textbf{\emph{\textcolor{DeepPurple}{Expected Output (excerpt)}}}

{\small
\begin{verbatim}
[
  {"page_number": 1, "word_count": 312, "line_count": 28},
  {"page_number": 2, "word_count": 445, "line_count": 35},
  ...
]
\end{verbatim}
}

Without the skill, agents commonly miscount words by splitting text on whitespace or miscount lines by using layout-preserved extraction, producing structurally valid but numerically incorrect outputs.
\end{tcolorbox}

\section{Limitations}
\label{sec:limitations}

SkillGenBench is intended as a controlled benchmark for skill-generation pipelines, but it does not cover every deployment setting for agent skills. First, the current dynamic execution results cover six generation backbones, and future releases should ship fully self-contained raw run directories for every summary row. 

Second, the completed-failure taxonomy is diagnostic rather than a substitute for human adjudication. It combines execution traces, generated code, and task metadata to classify completed verifier failures under shared mechanisms; this makes large-scale analysis possible, but individual failures can involve multiple overlapping causes.

Third, the benchmark focuses on deterministic task verifiers and fixed downstream execution. This is useful for isolating skill generation, but it under-represents settings where downstream agents can negotiate with users, call external services interactively, or revise skills after deployment. Fourth, the repository and document sources are broad enough to expose distinct failure modes, but they are not exhaustive. Additional domains, larger repositories, multi-repository workflows, and longer task-agnostic skill-library settings would further test whether generated skills transfer across related tasks. Finally, the static scores are rule-based proxies. They are useful for explaining observed failures, but they should be interpreted as diagnostics rather than intrinsic measures of skill quality.

\paragraph{Broader Impact.}
This work introduces a benchmark for evaluating skill generation in large language models, which may improve the reliability of agent systems. At the same time, enhanced automation capabilities may introduce risks such as misuse of generated workflows or lowering the barrier to executing complex tasks. Careful evaluation and monitoring are important to mitigate these risks.

\section{Benchmark Construction Details}
\label{app:construction_details}

\subsection{Human Verification}
\label{app:human_verification}

Automatic task verification provides an initial difficulty and reliability screen for candidate items. It removes candidates that are too easy, too brittle, or unlikely to yield stable verification under the intended source-access setting. Candidate tasks that pass these checks still require manual review for clarity, coverage, and alignment with the intended procedural-recovery problem. We therefore include a manual verification pass during benchmark construction. The audit is applied to each candidate task, including its source materials, task specification, test cases, verifier, and exposed skill materials. Its purpose is to check whether the item is clear, appropriately challenging, and quantitatively evaluable under the intended procedural recovery problem. The audit follows five criteria shown in Table~\ref{tab:human_verification_criteria}.

\begin{table}[h]
  \caption{Manual verification criteria for candidate benchmark tasks. A candidate is retained only when it satisfies all five criteria.}
  \label{tab:human_verification_criteria}
  \centering
  \small
  \setlength{\tabcolsep}{5pt}
  \renewcommand{\arraystretch}{1.10}
  \begin{tabular}{p{0.22\linewidth}p{0.38\linewidth}p{0.30\linewidth}}
    \toprule
    \textbf{Criterion} & \textbf{Acceptance criterion} & \textbf{Typical failure mode} \\
    \midrule
    Moderate difficulty & Requires nontrivial use of the source materials while remaining solvable from the provided context & Too easy, unrealistic, ambiguous, or underspecified \\
    Clear specification & States the expected inputs, outputs, constraints, and boundary conditions & Missing schema, vague target behavior, or conflicting requirements \\
    Quantitative evaluability & Can be judged by executable tests, structured checks, or well-defined artifact metrics & Requires open-ended subjective judgment \\
    Sufficient coverage & Test cases cover normal, edge, and adversarial conditions aligned with the task & Only covers the happy path or misses key constraints \\
    Test-case quality & References, tests, and exposed materials are consistent and do not leak answers or verifier-specific hints & Answer leakage, brittle checks, or inconsistent expected outputs \\
    \bottomrule
  \end{tabular}
\end{table}

During benchmark construction, we manually inspected 678 candidate tasks produced by the generation pipeline and retained 187 that satisfied all five criteria, corresponding to an acceptance rate of 27.6\%. Candidate tasks that failed the audit were revised and returned to the refinement loop when the issue was repairable, and discarded otherwise. This manual pass complements the automatic validation stage: automatic checks filter candidates by empirical solvability and verifier stability, while manual verification controls task clarity, evaluation coverage, and exposure leakage.


\definecolor{promptframe}{RGB}{55,65,90}
\definecolor{promptback}{RGB}{249,249,251}
\definecolor{prompttitle}{RGB}{255,255,255}

\newtcolorbox{promptbox}[1][]{
  enhanced jigsaw, breakable,
  colback=promptback, colframe=promptframe,
  boxrule=0.4pt, arc=2pt,
  left=6pt, right=6pt, top=4pt, bottom=4pt,
  fonttitle=\bfseries\small, coltitle=prompttitle,
  colbacktitle=promptframe,
  title={#1},
  before skip=6pt, after skip=8pt,
}

\lstdefinestyle{promptstyle}{
  basicstyle=\ttfamily\footnotesize,
  breaklines=true, breakatwhitespace=false,
  columns=fullflexible, keepspaces=true,
  showstringspaces=false, upquote=true,
  aboveskip=2pt, belowskip=2pt,
}


\subsection{Generation Prompts}
\label{app:prompts:gen}

This section lists the prompts driving each stage of our generation
pipeline. We follow Python's \texttt{str.format} convention: single
braces \texttt{\{var\}} mark runtime substitutions (e.g., the source
document, the KG summary, the slot index), while doubled braces
\texttt{\{\{...\}\}} are literal braces forwarded to the model, which is used
primarily inside the JSON schemas embedded in each prompt.

\subsubsection{Stage 1 --- Knowledge Graph Construction}

We extract a typed knowledge graph from each source document or code repository, then
detect communities and summarize them into theme-level descriptors that
serve as scenario-generation context downstream.

\paragraph{KG construction.}
A single-pass prompt that proposes entity types, extracts entities, and
emits subject--predicate--object triples in one structured JSON object.

\begin{promptbox}[Prompt 1: KG Construction]
\begin{lstlisting}[style=promptstyle]
## Task

Analyze the following text and perform three steps IN ORDER, producing a single
JSON output that contains the results of all three steps.

### Step 1: Propose Entity Types

Identify the entity types present in this text that are relevant to understanding
its content. The goal is to capture entities that have meaningful relationships
with other entities.

Rules:
- Avoid overly generic types like "other" or "unknown"
- Do NOT generate redundant or overlapping types (e.g., if the text has both
"company" and "organization", pick only one)
- Quality over quantity -- every type must be relevant

### Step 2: Extract Entities

Using the entity types you proposed in Step 1, extract all important entities
from the text.

Rules:
- Focus on substantive mentions -- entities central to the text or mentioned
with meaningful detail
- Prioritize entities with relationship potential -- they should connect to
other entities in the text
- Use the most widely recognized or official name as it appears in the text
- Use full names for people (e.g., "Marie Curie" not just "Curie")
- Normalize variants to one canonical form (e.g., "U.S." -> "United States")
- Each unique entity appears only once
- Be selective: a focused set of well-connected entities is better than a
comprehensive list with many isolates
- For each entity, provide:
  - id: a unique slug-style identifier (lowercase, underscores, e.g., "bigfoot", "myth_track")
  - name: canonical name
  - type: one of the types proposed in Step 1
  - description: concise description (1-3 sentences) based on what the text says

### Step 3: Extract Relations

Extract subject-predicate-object triples between the entities identified in Step 2.

Rules:
- **Subject** and **object** must both be entities from Step 2 (use their IDs)
- **Predicate** should be clear and specific (e.g., "founded_by", "located_in",
"part_of", "interacts_with", "scored_by")
- Avoid vague predicates like "related_to" or "associated_with"
- Only extract relationships explicitly stated or clearly implied in the text
- Ensure correct directionality (subject -> predicate -> object)
- Be thorough: maximize connectivity, minimize isolated entities
- For each relation provide a brief source_text quote supporting it

## Output Format

Return a single JSON object with this exact structure:
```json
{{
  "entity_types": ["type1", "type2", ...],
  "entities": [
    {{
      "id": "entity_id",
      "name": "Entity Name",
      "type": "entity_type",
      "description": "What this entity is based on the text"
    }}
  ],
  "relations": [
    {{
      "subject_id": "entity_id_1",
      "predicate": "relationship_verb",
      "object_id": "entity_id_2",
      "source_text": "brief quote from text supporting this relation"
    }}
  ]
}}
```

## Text to Analyze

{text}

## Extraction

Analyze the text above and return the JSON output. Be thorough but precise.
\end{lstlisting}
\end{promptbox}

\paragraph{Community summary.}
After running community detection on the merged KG, every community is
condensed into a short thematic summary that anchors downstream
scenario generation.

\begin{promptbox}[Prompt 2: Community Summary]
\begin{lstlisting}[style=promptstyle]
## Task

Summarize the following group of related entities and relationships from a document in 2-3 sentences.

Focus on what topic or theme this group covers.

## Entities ({entity_count})

{entity_lines_joined}
\end{lstlisting}
\end{promptbox}

\subsubsection{Stage 2 --- Scenario Generation}

Conditioning on the KG summary together with the source document, we
ask the model to propose practical, multi-section, computation-bearing
application scenarios that motivate the downstream tasks.

\begin{promptbox}[Prompt 3: Scenario Generation]
\begin{lstlisting}[style=promptstyle]
# Task Scenario Generation

You are a benchmark designer. Read the document below and propose **{num_scenarios}** realistic, practical application scenarios.

## Requirements

Each scenario must be a **real-world tool or system** that a human would actually build using this document's knowledge, such as:
- Rule engines, scoring systems, decision tools
- Automated assistants or advisory bots
- Process automation, workflow validators
- Data processing pipelines, calculators
- Compliance checkers, audit tools

### What makes a GOOD scenario:
- Requires integrating knowledge from **multiple sections** of the document
- Involves **computation, logic, or decision-making** -- not just information retrieval
- Would be genuinely useful to someone working in this domain
- Requires document-specific rules/data that cannot be guessed from common knowledge

### What to AVOID:
- Exam-style questions ("What is X?" or "List the Y")
- Simple fact lookups or single-step reasoning
- Scenarios where common knowledge suffices without the document
- Overly narrow scenarios that only touch one sentence in the document
- Scenarios that are too abstract or vague to implement

## Output Format

Output a JSON array of {num_scenarios} scenario objects:

```json
[
  {{
    "scenario_id": "scenario_001",
    "title": "Short descriptive title",
    "description": "2-3 sentences describing what the tool/system does",
    "real_world_use_case": "Who would use this and why",
    "applicable_rules": ["Rule/section 1 from doc", "Rule/section 2"],
    "domain_concepts": ["concept1", "concept2"],
    "data_needed": ["What specific data from the document is needed"],
    "problem_type_suggestion": "MC|FIB|CS|RANK|CODE",
    "data_strategy_suggestion": "doc_lookup|doc_logic|synthetic_scenario",
    "difficulty_estimate": "medium|hard",
    "cross_section_refs": ["Section A", "Section B"]
  }}
]
```

## Knowledge Graph Summary

{kg_summary}

## Document

{document}
\end{lstlisting}
\end{promptbox}

\subsubsection{Stage 3 --- Task and Test-Case Generation}

For every scenario slot, this prompt jointly produces (i)~a
\emph{function-interface} task description that abstracts away
document-specific constants and (ii)~an executable test-case bundle
whose \texttt{solve} function hardcodes those constants internally,
enforcing the contamination boundary central to our benchmark.

\begin{promptbox}[Prompt 4: Task / Test-Case Generation]
\begin{lstlisting}[style=promptstyle]
# Function-Based Task & Test Case Generation

Generate ONE task and ONE testcase bundle for the given scenario.

## Scenario
{scenario}

## Task Slot: {slot_idx}

## Full Knowledge Graph
{kg_summary}

## KG Chain (focused subset for this task)
{kg_chain}

## KEY CONCEPT: Function-Based Tasks

Every task describes a **general tool/function interface** -- NOT a specific one-off question with fixed numbers.

- The QUESTION describes WHAT the function does, its INPUT schema, and its OUTPUT schema.
- The QUESTION references document rules in abstract terms (e.g., "according to the document's threshold", "as specified in the procedure").
- The TEST CASES provide {test_cases_per_task} **different concrete scenarios** as input->output pairs.
- The SOLUTION CODE defines a `solve(input_data)` function with document-specific constants hardcoded inside.

### WRONG (too specific -- all numbers baked in):
```
"question": "A scheduler validates two Tuesday-cycle threads. Thread A has walls [5,12,23] with ages [3,8,15] and threshold 10..."
```
This bakes ALL specifics into the question. Only one answer is possible, and test cases can only check variable names.

### CORRECT (function interface):
```
"question": "Implement `solve(input_data)` as a **Wall Scheduler**.\n\nINPUT: dict with 'threads' (list of thread dicts with 'walls', 'max_age_threshold') and 'mandatory_walls' (list of IDs)\n\nOUTPUT: dict with 'selected_walls' per thread and 'halted' flag\n\nRULES:\n- Threshold is per-policy-excerpt (not global)\n- Noncompliance alert has selection precedence over age\n- Mandatory walls always included"
```
Test cases provide {test_cases_per_task} different inputs with different configurations and expected outputs.

## CRITICAL RULES

### Anti-Contamination
1. The **question text** describes the function interface using ABSTRACT references to document rules.
2. All **document-specific constants, thresholds, formulas, parameter values** are hardcoded INSIDE the `solve()` function body -- NOT in the question text, NOT in the test case inputs.
3. Test case inputs contain varying scenarios; the function applies document-internal knowledge to compute outputs.
4. The task must be IMPOSSIBLE to implement correctly without reading the document.

### Question Quality
5. Questions must require **multi-step reasoning** using multiple document rules.
6. Each question integrates information from **multiple parts** of the document.
7. The function should test **consequences and interactions** of document rules.
8. Input schema should be rich enough to support {test_cases_per_task} diverse test scenarios.

### Test Case Requirements
9. Generate exactly {test_cases_per_task} test cases with diverse scenarios.
10. Cover: normal cases, edge cases, boundary conditions, rule interactions.
11. Each test case: `{{"input": {{...}}, "expected_output": {{...}}}}`.
12. `solution_code` must define a function named `solve` taking one dict argument.
13. All document-specific values hardcoded inside `solve`.
14. `solve` must be self-contained -- no external variable references.
15. The `solve` function body should include comments citing which document section each constant comes from.

### Input Constant Consistency
16. All string constants in test case inputs (e.g., entity names, type names, category labels) MUST exactly match the strings used in `solve()` comparisons. If `solve()` checks `name == "Setting Manager"`, the test input must use `"Setting Manager"`, NOT `"SettingManager"` or `"setting_manager"`. Double-check: every string in input that will be compared inside `solve()` must be copy-pasted from the corresponding `solve()` code.

### Output Simplicity Constraints
17. The function output MUST be a FLAT or SHALLOW dict (max 2 nesting levels). No deeply nested reports, audits, or multi-section compilations.
18. Each test case expected_output MUST have at most 15 leaf values (strings, numbers, booleans, nulls). Count every terminal value in the nested structure. If your design exceeds 15, simplify the output schema.
19. The task MUST be a SINGLE DECISION or SINGLE COMPUTATION -- not "compile a full report". Good: "decide which walls to reset", "classify the input", "compute a score". Bad: "generate a comprehensive audit report", "compile a multi-section learning pack".
20. String values in expected_output SHOULD be short enums or codes (e.g., "PASS", "FAIL", "ALERT"), NOT long narrative sentences. If a reason/message field is needed, keep it under 50 characters.
21. The question text MUST be COMPLETE -- do not truncate input schemas or rule descriptions. If the schema is too large, simplify the function interface to fewer input fields. Every field mentioned in the INPUT section must have its type and description fully specified.

## Output Format

Output exactly TWO JSON blocks (```json fenced):

### Block 1: Task
```json
{{
  "task_id": "descriptive_tool_name_{slot_idx:03d}",
  "type": "tool_type",
  "question": "Implement a function `solve(input_data)` that acts as a **Tool Name**.\n\nINPUT: a dict with keys:\n- 'key1': type -- description\n- 'key2': type -- description\n\nOUTPUT: a dict with keys:\n- 'result_key': type -- description\n\nRULES (from the document):\n- Rule 1 (abstract, no specific values)\n- Rule 2 (abstract, no specific values)",
  "expected_output": {{
    "format": "dict",
    "key_results": {{}}
  }},
  "info_locations": [{{"name": "...", "location": "...", "description": "..."}}],
  "domain_knowledge_needed": [{{"knowledge": "...", "in_document": true}}],
  "reasoning_steps": ["step1", "step2"],
  "computation_chain_length": 5,
  "anti_contamination": {{
    "why_not_pretrain": "...",
    "why_skill_helps": "..."
  }}
}}
```

### Block 2: Testcase Bundle
```json
{{
  "task_id": "same_as_above",
  "test_id": "tc_{slot_idx:03d}",
  "setup_code": "import math\nimport numpy as np",
  "function_name": "solve",
  "solution_code": "def solve(input_data):\n    # Extract input\n    key1 = input_data['key1']\n    key2 = input_data['key2']\n    # Document-specific constants (from Section X.Y)\n    THRESHOLD = 42  # Section 3.2\n    RATE = 0.15  # Table 1\n    # Apply document rules\n    ...\n    return {{'result_key': computed_value}}",
  "test_cases": [
    {{
      "input": {{"key1": "scenario_A_value", "key2": 10}},
      "expected_output": {{"result_key": "expected_A"}}
    }},
    {{
      "input": {{"key1": "scenario_B_value", "key2": 20}},
      "expected_output": {{"result_key": "expected_B"}}
    }},
    {{
      "input": {{"key1": "edge_case", "key2": 0}},
      "expected_output": {{"result_key": "expected_edge"}}
    }}
  ],
  "timeout_seconds": 120
}}
```

## Difficulty Amplification
- The `solve()` function MUST hardcode at least 3 distinct document-specific constants from different sections.
- Rules MUST come from at least 2 different document topics/sections.
- Include at least 1 edge case where a naive interpretation gives a WRONG answer.
- Include at least 1 rule interaction that produces a counter-intuitive result.
- At least 2 of the {test_cases_per_task} test cases should trigger edge-case behavior.
- Constants should NOT be round numbers or common values -- use the exact values from the document.

## Document

{document}
\end{lstlisting}
\end{promptbox}

\subsubsection{Stage 4 --- Validation and Refinement}

Each candidate task is screened along three orthogonal axes before
acceptance: an LLM judge rates eight quality dimensions; a
\emph{corpus-free} solver attempt estimates pretrain-contamination
risk; and a \emph{with-corpus} solver attempt verifies that the
document is actually sufficient. Failing tasks are routed back through
a refinement prompt rather than discarded.

\paragraph{Multi-dimensional verification.}\mbox{}\\[1pt]

\begin{promptbox}[Prompt 5: Task Verification]
\begin{lstlisting}[style=promptstyle]
# Multi-Dimensional Question Evaluation

Evaluate the quality of a benchmark question based on the evidence below.

## Evidence
{evidence}

## Evaluation Criteria

Rate each dimension on a 0.0-1.0 scale:

1. **complexity_score**: How many reasoning steps are required? Simple lookup = 0.2, multi-step computation = 0.8+
2. **utility_score**: How practical is this task? Would someone actually need to solve it? Exam-style = 0.2, real tool = 0.8+
3. **contamination_risk**: Based on pretrain_score, how likely is pretrain contamination? >20% pretrain = high risk
4. **doc_dependency**: How much does the answer depend on document-specific info? Generic knowledge suffices = 0.2, fully doc-dependent = 1.0
5. **skill_benefit**: How much would a good skill document help? No help = 0.2, critical = 1.0
6. **doc_only_sufficient**: Based on doc_only_score, can the task be solved with the document? <10% = critically too hard (needs simplification). 10-40% = ideal range. >50% = may be too easy.
7. **doc_adds_value**: Does the document provide advantage over pretrain alone? doc_only_score - pretrain_score < 10% = document not helping.
8. **output_testability**: Are the expected outputs structural and objectively verifiable?
   - Outputs dominated by long freeform strings (messages, recommendations, dialog >30 chars) test string reproduction rather than document knowledge -> score 0.0-0.2
   - Outputs that are numbers, booleans, short enums, computed values -> score 0.8+
   - Outputs with emoji characters -> score 0.0
   - Deeply nested outputs with >20 leaf values per test case -> score 0.2 (too complex to match)
   - If >70% of test cases produce identical output -> score 0.2 (lacks discriminability)

## Also provide:
- **verdict**: "pass" or "fail"
- **fail_reasons**: list of specific reasons if failing
- **improvement_suggestions**: specific ways to make the question harder/better

## Output Format

```json
{{
  "complexity_score": 0.7,
  "utility_score": 0.8,
  "contamination_risk": "low|medium|high",
  "doc_dependency": 0.9,
  "skill_benefit": 0.7,
  "output_testability": 0.8,
  "verdict": "pass",
  "fail_reasons": [],
  "improvement_suggestions": ["suggestion 1"]
}}
```
\end{lstlisting}
\end{promptbox}

\paragraph{Corpus-free solvability check.}
The candidate solver receives only the task statement; the resulting
pass-rate estimates how much of the answer is recoverable from
parametric knowledge alone (the contamination floor).

\begin{promptbox}[Prompt 6: Corpus-Free Solvability Check]
\begin{lstlisting}[style=promptstyle]
Task: {question}

Write a Python function named `solve` that implements the tool described above.
The function takes a single dict argument and returns the output as described.

Requirements:
1. Import any necessary libraries at the top level
2. Define the `solve(input_data)` function
3. Hardcode all necessary constants inside the function
4. Return the result as specified in the task

Return ONLY executable Python code. No markdown fences, no explanations.
\end{lstlisting}
\end{promptbox}

\paragraph{With-corpus triviality check.}
The same solver is rerun with the source document attached;
\(\text{doc\_only} - \text{pretrain}\) quantifies how much value the
document actually contributes.

\begin{promptbox}[Prompt 7: With-Corpus Triviality Check]
\begin{lstlisting}[style=promptstyle]
You are given the following reference document:

{document}

---

Task: {question}

Write a Python function named `solve` that implements the tool described above.
The function takes a single dict argument and returns the output as described.

Requirements:
1. Import any necessary libraries at the top level
2. Define the `solve(input_data)` function
3. Hardcode all necessary constants (from the document) inside the function
4. Return the result as specified in the task

Return ONLY executable Python code. No markdown fences, no explanations.
\end{lstlisting}
\end{promptbox}

\paragraph{Targeted refinement.}
Rather than discarding rejected tasks, the verifier's failure reasons
are forwarded to a refinement prompt that surgically fixes the
identified issue (contamination, over-difficulty, string-matching
output, low diversity, etc.) while preserving \texttt{task\_id} and
\texttt{test\_id}.

\begin{promptbox}[Prompt 8: Task / Test-Case Refinement]
\begin{lstlisting}[style=promptstyle]
# Question Improvement

The question below failed verification. Improve it based on the analysis.

## Original Task
{task}

## Original Testcase Bundle
{testcase}

## Verification Analysis
{analysis}

## Document Excerpt (for reference)
{document_excerpt}

## Improvement Instructions

Based on the failure analysis, improve the question and testcase:

1. **If pretrain contaminated**: Make the question more document-specific. The `solve` function should require document-specific constants/rules that cannot be guessed. Reference obscure details, combine multiple rules, or require document-specific parameter values hardcoded in the function body.
2. **If too simple**: Add more computation steps, require cross-referencing multiple sections, or ask about consequences rather than facts.
3. **If low utility**: Reframe as a practical tool or system that someone would actually build.
4. **If solution code broken**: Fix the `solve` function while keeping the question intent.
5. **If too hard (doc_only too low)**: The task is too complex for the LLM to solve even with the full document. Simplify by: reducing the number of rules/steps required, making the expected output format simpler (fewer nested keys), breaking the task into a smaller more focused scope, using more standard output types (single value, simple dict) instead of complex nested structures. Keep the core document-dependency but reduce implementation complexity.
6. **If doc_not_helping (doc_only =~ pretrain)**: The document doesn't provide meaningful advantage. Make the task more document-specific by: requiring document-specific constants that cannot be guessed, referencing obscure details or unique terminology from the document, combining rules from multiple non-obvious sections of the document.
7. **If string_matching_dominant**: The task's expected outputs rely on exact matching of long freeform strings (>30 chars). This tests string reproduction, NOT document knowledge. Restructure the output to test STRUCTURAL decisions:
   - Messages/replies -> action codes + rule IDs: `{{"reply": "Hold on!..."}}` -> `{{"action": "SAFETY_BLOCK", "rule_triggered": "R3_ELECTRICAL", "safety_level": "HIGH"}}`
   - Recommendations -> category enums + scores: `{{"advice": "Passage p1: paraphrase..."}}` -> `{{"risk_level": "HIGH", "primary_issue": "SIMILARITY_EXCEEDED", "exceeded_by": 1.0}}`
   - Status messages -> boolean flags + enums: `{{"message": "You are eligible!"}}` -> `{{"eligible": true, "action": "PROCEED", "redirect_target": null}}`
   - Descriptions -> structured facts: `{{"issue": "Absolute claim not allowed for 'Joker'..."}}` -> `{{"issue_type": "ABSOLUTE_CLAIM", "theme_keyword": "Joker", "violating_words": ["inherently"]}}`
   - All strings must be SHORT (<30 chars) from a CLOSED SET. NEVER use emoji. Add numeric/boolean outputs. Preserve the SAME document knowledge requirement.
8. **If output_overly_complex**: The task's expected output is too deeply nested (>20 leaf values per test case), causing format-sensitive failures even when the logic is correct. Simplify the output structure:
   - Flatten nested dicts: instead of `{{"responses": [{{"task_id": "...", "status": "...", "new_totals": {{"total_spent": ..., "by_category": {{...}}}}}}]}}`, return a simpler summary: `{{"total_spent": ..., "last_action_status": "success", "rejected_count": 0}}`
   - Reduce the number of output fields to the ESSENTIAL ones that test document knowledge
   - Remove intermediate/debugging fields (detailed per-command responses, full category breakdowns)
   - Keep at most 2 levels of nesting in the output dict
   - Focus on the FINAL computed result, not the step-by-step trace
9. **If emoji_in_output**: Remove ALL emoji characters from expected output values. Replace emoji-containing messages with structured action codes or boolean flags. Emoji make exact matching impossible for LLMs and do not test document knowledge.
10. **If low_output_diversity**: The task produces nearly identical output for most test cases, meaning it lacks discriminability -- the LLM could score high just by returning a fixed template. Fix by:
    - Adding more varied input scenarios that trigger DIFFERENT code paths and produce DIFFERENT outputs
    - Ensuring at least 50% of test cases have structurally different expected outputs
    - Including edge cases and boundary conditions that produce distinct results
    - Each test case should exercise a different rule or combination of rules from the document

## CRITICAL RULES -- Function-Based Format
- Keep the same task_id and test_id
- The question describes a GENERAL function interface (input schema -> rules -> output schema)
- The question must NOT contain document-specific values/numbers
- `solution_code` must define a `solve(input_data)` function
- All document-specific constants hardcoded INSIDE the `solve` function
- Include 5-10 diverse test cases as input->output pairs
- Each test case tests a different scenario/edge case

## Output Format

Output exactly TWO JSON blocks (```json fenced): one improved task, one improved testcase bundle.
Use the same schema as the originals but with the function-based format.
\end{lstlisting}
\end{promptbox}

\subsection{Evaluation Prompts}
\label{app:prompts:eval}
 
\begin{promptbox}[Prompt 9: Agent Evaluation Instruction]
\begin{lstlisting}[style=promptstyle]
Read /workspace/instruction.md for the full task description.
Input files are at /workspace/input/.
Documentation is at /workspace/docs/.
Write the complete Python solution to /workspace/agent_output.py. The file must be importable and define any required functions (e.g. solve()).
IMPORTANT: You MUST use the Skill tool at least once before implementing to load and follow the most relevant skill guidance.
If the triggered skill references bundled helper files such as scripts/ or references/, use the mirrored skill package under /workspace/skill/.
\end{lstlisting}
\end{promptbox}

%

\end{document}